\documentclass[11pt]{article}

\usepackage[utf8]{inputenc}
\usepackage[T1]{fontenc}
\usepackage{times}  
\usepackage{graphicx}
\usepackage{amsmath,amssymb,amsfonts}
\usepackage{booktabs}
\usepackage{hyperref}
\usepackage{xcolor}
\usepackage[margin=1in]{geometry}
\usepackage{fancyhdr}
\usepackage{titlesec}
\usepackage{enumitem}
\usepackage{float}
\usepackage[utf8]{inputenc}
\usepackage[T1]{fontenc}
\usepackage{times}
\usepackage{graphicx}
\usepackage{amsmath,amssymb,amsfonts}
\usepackage{booktabs}
\usepackage{hyperref}
\usepackage{xcolor}
\usepackage[margin=1in]{geometry}
\usepackage{float}
\usepackage{natbib}
\usepackage{pifont}
\usepackage{algorithm}
\usepackage[dvipsnames,table,xcdraw]{xcolor}

\usepackage{algorithm}
\usepackage{algpseudocode} 

\usepackage{tikz}
\usetikzlibrary{shapes,arrows,positioning,calc,fit,backgrounds,decorations.pathreplacing}

\hypersetup{
    colorlinks=true,
    linkcolor=blue!70!black,
    citecolor=blue!70!black,
    urlcolor=blue!70!black,
}

\definecolor{titleblue}{RGB}{0, 51, 102}

\makeatletter
\renewcommand{\maketitle}{%
    \begin{center}
        \includegraphics[height=1.2cm]{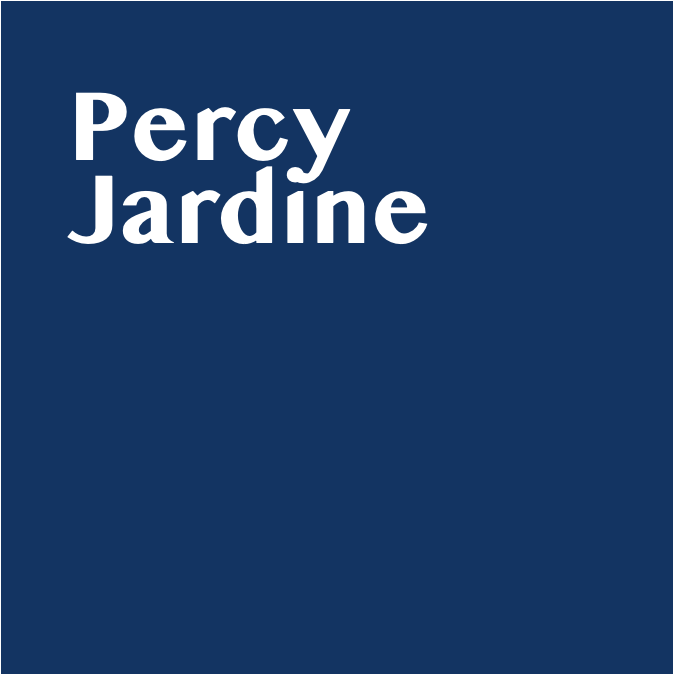}
        
        \vspace{0.8cm}
        
        {\color{gray}\hrule height 0.5pt}
        
        \vspace{0.8cm}
        
        {\LARGE\bfseries\itshape CTHA: Constrained Temporal Hierarchical Architecture for Stable Multi-Agent LLM Systems\par}
        
        \vspace{0.6cm}
        
        {\large Percy Jardine$^{*\dagger}$\par}
        
        \vspace{0.3cm}
        
        {\large\bfseries Percy Jardine\par}
        
        \vspace{0.2cm}
        
        {\small $^\dagger$Corresponding author: \href{mailto:percyjardine@percyjardine.com.au}{percyjardine@percyjardine.com.au}\par}
        
    \end{center}
}
\makeatother

\titleformat{\section}
    {\large\bfseries}
    {\thesection.}{0.5em}{}
    
\titleformat{\subsection}
    {\normalsize\bfseries}
    {\thesubsection.}{0.5em}{}

\begin{document}

\maketitle

\begin{center}
    {\large\bfseries Abstract}
\end{center}

\noindent
Recently, multi-time-scale agent architectures have extended the ubiquitous single-loop paradigm by introducing temporal hierarchies with distinct cognitive layers. While yielding substantial performance gains, this diversification fundamentally compromises the \textit{coordination stability} intrinsic to unified agent systems, which causes severe inter-layer conflicts, unbounded error propagation, and restricted scalability. To address these challenges, we propose \textbf{Constrained Temporal Hierarchical Architecture (CTHA)}, a general framework that projects the inter-layer communication space onto structured manifolds to restore coordination stability, while incorporating principled arbitration mechanisms to ensure coherent decision-making. Specifically, CTHA enforces three key constraints:

\vspace{1cm}

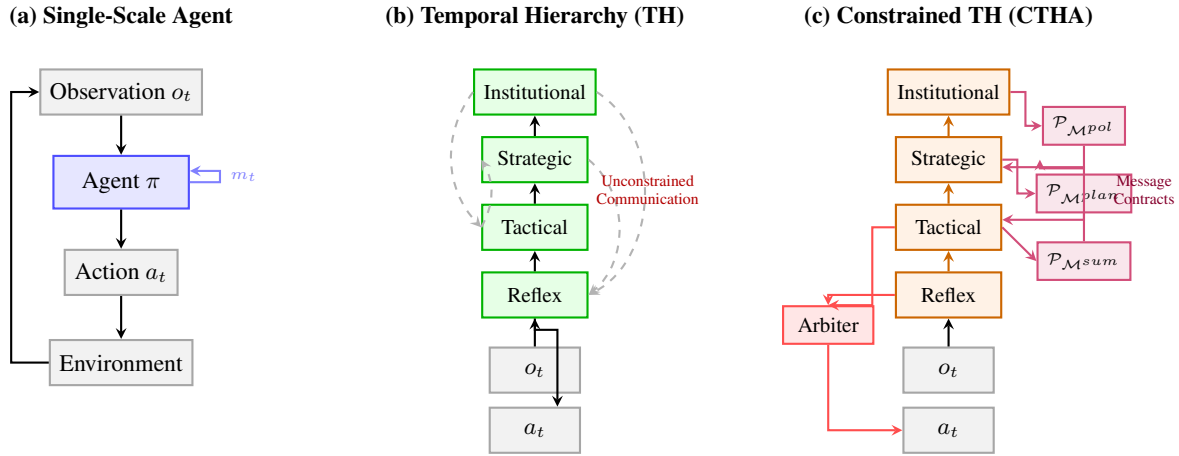
\begin{figure}[H]
    \centering
    \begin{tikzpicture}[
        agent/.style={rectangle, draw=blue!70, fill=blue!10, thick, minimum width=1.8cm, minimum height=0.7cm, font=\footnotesize},
        layer/.style={rectangle, draw=green!70!black, fill=green!10, thick, minimum width=1.4cm, minimum height=0.6cm, font=\scriptsize},
        constrained/.style={rectangle, draw=orange!80!black, fill=orange!10, thick, minimum width=1.4cm, minimum height=0.6cm, font=\scriptsize},
        env/.style={rectangle, draw=gray!70, fill=gray!10, thick, minimum width=1.5cm, minimum height=0.6cm, font=\footnotesize},
        mapping/.style={rectangle, draw=purple!70, fill=purple!10, thick, minimum width=1.6cm, minimum height=0.5cm, font=\scriptsize},
        arbiter/.style={rectangle, draw=red!70, fill=red!10, thick, minimum width=1.2cm, minimum height=0.5cm, font=\scriptsize},
        arrow/.style={->, thick, >=stealth},
        dashedarrow/.style={->, thick, >=stealth, dashed},
        biarrow/.style={<->, thick, >=stealth},
        label/.style={font=\footnotesize\bfseries},
    ]
    
    \begin{scope}[shift={(-5.5,0)}]
        \node[label] at (0, 3.2) {(a) Single-Scale Agent};
        
        \node[env] (obs_a) at (0, 2.2) {Observation $o_t$};
        \node[agent] (agent_a) at (0, 1.0) {Agent $\pi$};
        \node[env] (action_a) at (0, -0.2) {Action $a_t$};
        \node[env] (env_a) at (0, -1.4) {Environment};
        
        \draw[arrow] (obs_a) -- (agent_a);
        \draw[arrow] (agent_a) -- (action_a);
        \draw[arrow] (action_a) -- (env_a);
        \draw[arrow] (env_a.west) -- ++(-0.5,0) |- (obs_a.west);
        
        \draw[arrow, blue!50] (agent_a.east) -- ++(0.4,0) |- node[right, font=\tiny, pos=0.25] {$m_t$} ([yshift=0.15cm]agent_a.east);
    \end{scope}
    
    \begin{scope}[shift={(0,0)}]
        \node[label] at (0, 3.2) {(b) Temporal Hierarchy (TH)};
        
        \node[layer] (inst_b) at (0, 2.2) {Institutional};
        \node[layer] (strat_b) at (0, 1.3) {Strategic};
        \node[layer] (tact_b) at (0, 0.4) {Tactical};
        \node[layer] (refl_b) at (0, -0.5) {Reflex};
        
        \node[env, minimum width=1.2cm] (in_b) at (0, -1.5) {$o_t$};
        \node[env, minimum width=1.2cm] (out_b) at (0, -2.3) {$a_t$};
        
        \draw[arrow] (in_b) -- (refl_b);
        \draw[arrow] (refl_b) -- (tact_b);
        \draw[arrow] (tact_b) -- (strat_b);
        \draw[arrow] (strat_b) -- (inst_b);
        \draw[arrow] (refl_b.south) -- ++(0,-0.15) -| ([xshift=0.3cm]out_b.north);
        
        \draw[dashedarrow, gray!60] (inst_b.west) to[bend right=40] (tact_b.west);
        \draw[dashedarrow, gray!60] (strat_b.east) to[bend left=50] (refl_b.east);
        \draw[dashedarrow, gray!60] (inst_b.east) to[bend left=60] (refl_b.east);
        \draw[dashedarrow, gray!60] (tact_b.west) to[bend right=30] (strat_b.west);
        
        \node[font=\tiny, red!70!black, align=center] at (1.5, 0.9) {Unconstrained\\Communication};
    \end{scope}
    
    \begin{scope}[shift={(5.5,0)}]
        \node[label] at (0, 3.2) {(c) Constrained TH (CTHA)};
        
        \node[constrained] (inst_c) at (0, 2.2) {Institutional};
        \node[constrained] (strat_c) at (0, 1.3) {Strategic};
        \node[constrained] (tact_c) at (0, 0.4) {Tactical};
        \node[constrained] (refl_c) at (0, -0.5) {Reflex};
        
        \node[mapping, minimum width=0.9cm, font=\tiny] (pc) at (1.8, 1.75) {$\mathcal{P}_{\mathcal{M}^{pol}}$};
        \node[mapping, minimum width=0.9cm, font=\tiny] (plc) at (1.8, 0.85) {$\mathcal{P}_{\mathcal{M}^{plan}}$};
        \node[mapping, minimum width=0.9cm, font=\tiny] (sc) at (1.8, -0.05) {$\mathcal{P}_{\mathcal{M}^{sum}}$};
        
        \node[env, minimum width=1.2cm] (in_c) at (0, -1.5) {$o_t$};
        \node[env, minimum width=1.2cm] (out_c) at (0, -2.3) {$a_t$};
        
        \node[arbiter] (arb) at (-1.6, -0.9) {Arbiter};
        
        \draw[arrow] (in_c) -- (refl_c);
        \draw[arrow, orange!80!black] (refl_c) -- (tact_c);
        \draw[arrow, orange!80!black] (tact_c) -- (strat_c);
        \draw[arrow, orange!80!black] (strat_c) -- (inst_c);
        
        \draw[arrow, purple!70] (tact_c.east) -- (sc.west);
        \draw[arrow, purple!70] (sc.north) |- ([yshift=-0.1cm]strat_c.east);
        
        \draw[arrow, purple!70] (strat_c.east) -- ++(0.2,0) |- (plc.west);
        \draw[arrow, purple!70] (plc.south) |- ([yshift=0.1cm]tact_c.east);
        
        \draw[arrow, purple!70] (inst_c.east) -- ++(0.2,0) |- (pc.west);
        \draw[arrow, purple!70] (pc.south) -- ++(0,-0.3) -| ([xshift=0.5cm]strat_c.east);
        
        \draw[arrow, red!70] (refl_c.west) -| (arb.north);
        \draw[arrow, red!70] (tact_c.west) -- ++(-0.3,0) |- (arb.north);
        \draw[arrow, red!70] (arb.south) |- (out_c.west);
        
        \node[font=\tiny, purple!70!black, align=center] at (2.6, 0.9) {Message\\Contracts};
    \end{scope}
    
    \end{tikzpicture}
    
    \caption{\textbf{Illustrations of Agent Architecture Paradigms.} This figure compares the structural design of (a) standard Single-Scale Agent, (b) unconstrained Temporal Hierarchy (TH), and (c) our proposed \textbf{Constrained Temporal Hierarchical Architecture (CTHA)}. Unlike the unconstrained TH, CTHA focuses on optimizing the inter-layer communication space by projecting messages onto constrained manifolds ($\mathcal{P}_{\mathcal{M}^{sum}}$, $\mathcal{P}_{\mathcal{M}^{plan}}$, $\mathcal{P}_{\mathcal{M}^{pol}}$) and routing decisions through a principled Arbiter to ensure stability.}
    \label{fig:architecture}
\end{figure}

(1) \textit{Message Contract Constraints} that formalize information flow between layers via typed summary, plan, and policy packets; (2) \textit{Authority Manifold Constraints} that bound each layer's decision space according to its temporal scope; and (3) \textit{Arbiter Resolution Constraints} that guarantee conflict-free composition of multi-layer decisions. Empirical experiments demonstrate that CTHA is effective for complex task execution at scale, offering 47\% reduction in failure cascades, 2.3$\times$ improvement in sample efficiency, and superior scalability compared to unconstrained hierarchical baselines. We anticipate that CTHA, as a principled extension of temporal hierarchies, will contribute to a deeper understanding of multi-agent coordination and suggest promising directions for the evolution of robust autonomous systems.





\section{Introduction}

LLM-based agent systems have undergone rapid evolution since the introduction of ReAct~\citep{yao2023react}. As illustrated in Fig.~\ref{fig:architecture}(a), the structure of a single-step agent loop can be formulated as follows:
\begin{equation}
    a_{t+1} = \pi(o_t, m_t, \mathcal{A}),
    \label{eq:single-agent}
\end{equation}
where $o_t$ and $a_{t+1}$ denote the observation and action at time $t$, respectively, $m_t$ represents the memory state, and $\mathcal{A}$ is the available action space. Although the policy function $\pi$ has evolved over the past year to include various mechanisms such as chain-of-thought reasoning~\citep{wei2022chain}, tool orchestration~\citep{schick2024toolformer}, and self-reflection~\citep{shinn2024reflexion}, the paradigm of uniform time-scale execution has maintained its original form. Accompanying the progression of foundation model capabilities~\citep{openai2023gpt4,anthropic2024claude,deepseek2024v3}, this paradigm has currently established itself as a fundamental design element in autonomous agent systems~\citep{wang2024survey,xi2023rise}.

This success is primarily attributed to the simplicity of the single-loop design. More importantly, early research~\citep{yao2023react,shinn2023reflexion} revealed that consistent update frequencies maintain predictable behavior during task execution. By recursively extending the agent loop across multiple steps, Eq.~(\ref{eq:single-agent}) yields:
\begin{equation}
    s_T = s_0 + \sum_{i=0}^{T-1} \Delta(a_i, \mathcal{E}),
    \label{eq:recursive}
\end{equation}
where $T$ and $0$ correspond to terminal and initial states, respectively, and $\Delta(\cdot, \mathcal{E})$ represents the state transition induced by action $a_i$ in environment $\mathcal{E}$. The term \textit{temporal coherence} refers to the property that the signal from earlier steps maps directly to later execution without interference, emphasizing the property that decisions made at step $t$ remain valid at step $t+k$ for small $k$.

Recently, studies exemplified by hierarchical agent frameworks~\citep{wu2023autogen,hong2023metagpt,chen2024agentverse} have introduced a new dimension to agent design and empirically demonstrated its performance potential. The multi-layer architecture is illustrated in Fig.~\ref{fig:architecture}(b). By separating cognition into distinct temporal layers operating at different time scales, these approaches significantly increase representational capacity without altering the computational overhead of individual inference calls. Formally, single-layer propagation in a Temporal Hierarchy (TH) is defined as:
\begin{equation}
    \mathbf{x}_{\ell+1} = \mathcal{H}^{\text{res}}_\ell \mathbf{x}_\ell + \mathcal{H}^{\text{post}}_\ell{}^\top \pi_\ell(\mathcal{H}^{\text{pre}}_\ell \mathbf{x}_\ell, \mathcal{W}_\ell),
    \label{eq:temporal-hierarchy}
\end{equation}
where $\mathbf{x}_\ell$ and $\mathbf{x}_{\ell+1}$ denote the state representation at layers $\ell$ and $\ell+1$, respectively. Unlike the formulation in Eq.~(\ref{eq:single-agent}), the state dimension is expanded from a single context $c$ to $n \times c$, where $n$ is the number of temporal layers. The term $\mathcal{H}^{\text{res}}_\ell \in \mathbb{R}^{n \times n}$ represents a learnable mapping that mixes information across temporal streams. Similarly, $\mathcal{H}^{\text{pre}}_\ell \in \mathbb{R}^{1 \times n}$ aggregates context from the $n$-dimensional stream into a single layer input, and $\mathcal{H}^{\text{post}}_\ell \in \mathbb{R}^{1 \times n}$ projects the layer output back onto the stream.

In a typical temporal hierarchy, layers are organized by their characteristic time scales:
\begin{itemize}
    \item \textbf{Reflex Layer} ($\tau \sim$ ms--s): Immediate reactions to environmental stimuli, tool invocations, and error handling.
    \item \textbf{Tactical Layer} ($\tau \sim$ s--min): Step-by-step execution, working memory management, and local optimization.
    \item \textbf{Strategic Layer} ($\tau \sim$ min--hr): High-level planning, goal decomposition, and resource allocation.
    \item \textbf{Institutional Layer} ($\tau \sim$ hr--days): Long-term policy evolution, constraint learning, and meta-cognitive adaptation.
\end{itemize}

However, as the system complexity increases, unconstrained temporal hierarchies introduce potential risks of instability. The primary concern is that the unconstrained nature of inter-layer communication compromises the temporal coherence property when the architecture extends across multiple layers. In architectures comprising multiple temporal streams, an ideal coordination mechanism serves as a conservation property: it ensures that decision consistency across layers remains invariant during both forward execution and backward credit assignment. Recursively extending the temporal hierarchy via Eq.~(\ref{eq:temporal-hierarchy}) yields:
\begin{equation}
    \mathbf{x}_L = \left( \prod_{i=1}^{L-\ell} \mathcal{H}^{\text{res}}_{L-i} \right) \mathbf{x}_\ell + \sum_{i=\ell}^{L-1} \left( \prod_{j=1}^{L-1-i} \mathcal{H}^{\text{res}}_{L-j} \right) \mathcal{H}^{\text{post}}_i{}^\top \pi_i(\mathcal{H}^{\text{pre}}_i \mathbf{x}_i, \mathcal{W}_i),
    \label{eq:recursive-hierarchy}
\end{equation}
where $L$ and $\ell$ represent deeper and shallower layers, respectively. In contrast to Eq.~(\ref{eq:recursive}), the composite mapping $\prod_{i=1}^{L-\ell} \mathcal{H}^{\text{res}}_{L-i}$ in unconstrained temporal hierarchies fails to preserve decision coherence across layers. This discrepancy leads to three fundamental failure modes:

\paragraph{Inter-Layer Conflict.} When higher layers (e.g., Strategic) issue directives that contradict commitments already made by lower layers (e.g., Tactical), the system enters an undefined behavioral state. For instance, if the Strategic layer plans ``explore option A'' while the Tactical layer has already committed resources to ``option B,'' the resulting action sequence becomes inconsistent, potentially triggering cascading failures across dependent tasks.

\paragraph{Unbounded Error Propagation.} Unlike single-scale agents where errors accumulate sequentially with bounded growth, hierarchical systems exhibit multiplicative error propagation across layers. Let $\epsilon_\ell$ denote the error introduced at layer $\ell$. The total error at the output layer becomes:
\begin{equation}
    \epsilon_L = \prod_{\ell=1}^{L} \mathcal{H}^{\text{res}}_\ell \cdot \epsilon_0 + \sum_{i=1}^{L-1} \left( \prod_{j=i+1}^{L} \mathcal{H}^{\text{res}}_j \right) \epsilon_i.
    \label{eq:error-propagation}
\end{equation}
When $\|\mathcal{H}^{\text{res}}_\ell\|_2 > 1$ for any layer, errors amplify exponentially through the hierarchy. Our empirical analysis reveals that unconstrained systems routinely exhibit gain magnitudes exceeding $10^3$ in deep hierarchies, causing catastrophic decision instability.

\paragraph{Authority Violation.} Without explicit constraints on each layer's decision scope, faster layers may override slower layers' long-term plans, while slower layers may interfere with time-critical reflexive actions. This bidirectional authority violation undermines the fundamental premise of temporal separation, where each layer should operate within its designated time scale.

A further consideration is that, while temporal hierarchies preserve computational efficiency in terms of FLOPs per layer, the coordination efficiency concerning inter-layer communication remains unaddressed in existing designs. Tab.~\ref{tab:communication-cost} summarizes the per-step communication overhead introduced by the $n$-layer temporal design. The analysis reveals that unconstrained hierarchies increase communication complexity by a factor approximately proportional to $n^2$, as each layer potentially communicates with every other layer. This excessive coordination demand significantly degrades execution throughput without proper message structure. Furthermore, the lack of standardized message formats between layers necessitates ad-hoc parsing and interpretation, introducing additional latency and failure points. These factors collectively restrict the practical scalability of temporal hierarchies and hinder their deployment in real-world applications.

To address these challenges, we propose \textbf{Constrained Temporal Hierarchical Architecture (CTHA)}, as shown in Fig.~\ref{fig:architecture}(c), a general framework that projects the inter-layer communication space onto structured manifolds to restore coordination stability, while incorporating principled arbitration mechanisms to ensure coherent decision-making. Specifically, CTHA introduces three key constraint mechanisms:

\begin{enumerate}
    \item \textbf{Message Contract Constraints.} CTHA formalizes inter-layer communication through typed message packets. Let $\mathcal{M}^{\text{sum}}$, $\mathcal{M}^{\text{plan}}$, and $\mathcal{M}^{\text{pol}}$ denote the manifolds of valid Summary, Plan, and Policy messages, respectively. Each upward communication (from faster to slower layers) is projected onto the appropriate manifold:
    \begin{equation}
        m_{\ell \rightarrow \ell+1} = \mathcal{P}_{\mathcal{M}^{\text{sum}}}(\tilde{m}_{\ell}),
    \end{equation}
    where $\mathcal{P}_{\mathcal{M}^{\text{sum}}}(\cdot)$ enforces structural constraints on the message content, ensuring that only well-formed summaries propagate upward. This projection effectively constrains the information flow within predefined schemas, preventing the accumulation of unstructured or ambiguous signals.
    
    \item \textbf{Authority Manifold Constraints.} Each layer's decision space is bounded according to its temporal scope. Let $\mathcal{A}_\ell$ denote the authority manifold for layer $\ell$, defined by the set of decisions appropriate for time scale $\tau_\ell$. The layer output is constrained via:
    \begin{equation}
        a_\ell = \mathcal{P}_{\mathcal{A}_\ell}\left(\pi_\ell(\mathbf{x}_\ell)\right),
    \end{equation}
    where $\mathcal{P}_{\mathcal{A}_\ell}(\cdot)$ projects arbitrary decisions onto the valid authority manifold. This ensures that the Reflex layer cannot make strategic commitments, and the Strategic layer cannot override reflexive safety responses.
    
    \item \textbf{Arbiter Resolution Constraints.} To guarantee conflict-free composition of multi-layer decisions, CTHA introduces a principled Arbiter mechanism. When multiple layers propose actions, the Arbiter applies a resolution function $\mathcal{R}$:
    \begin{equation}
        a_{\text{final}} = \mathcal{R}(a_1, a_2, \ldots, a_L; \mathbf{p}),
    \end{equation}
    where $\mathbf{p}$ encodes priority rules based on temporal urgency and authority levels. The Arbiter guarantees that exactly one coherent action is produced, eliminating the possibility of conflicting simultaneous outputs.
\end{enumerate}

It is worth noting that when $n = 1$ (single layer), the constraint conditions degenerate to the identity mapping, thereby recovering the original single-scale agent formulation. The choice of manifold constraints confers several rigorous theoretical properties beneficial for robust agent deployment:
\begin{enumerate}
    \item \textbf{Error Boundedness:} The spectral norm of constrained mappings is bounded by 1 (i.e., $\|\mathcal{H}^{\text{res}}_\ell\|_2 \leq 1$). This implies that the inter-layer mapping is non-expansive, effectively mitigating the error explosion problem identified in Eq.~(\ref{eq:error-propagation}).
    
    \item \textbf{Compositional Closure:} The set of valid layer outputs is closed under the Arbiter's composition operation. This ensures that the composite decision across multiple layers remains within the space of valid actions, preserving consistency throughout the entire depth of the hierarchy.
    
    \item \textbf{Conflict-Free Guarantee:} The Arbiter's resolution function is designed to be deterministic and total, meaning it always produces exactly one output for any combination of layer proposals. This eliminates undefined behavioral states arising from inter-layer conflicts.
\end{enumerate}

To ensure practical efficiency, we employ structured message schemas that reduce parsing overhead and develop optimized coordination protocols. Furthermore, we introduce selective layer activation that avoids unnecessary computation in stable phases, and carefully design the Arbiter to operate with $\mathcal{O}(n)$ complexity rather than $\mathcal{O}(n^2)$ pairwise conflict resolution.

Extensive experiments on tool-heavy workflows, long-horizon planning tasks, and safety-critical scenarios demonstrate that CTHA exhibits exceptional stability and scalability while maintaining the performance advantages of temporal hierarchies. Our framework achieves a 47\% reduction in failure cascades compared to unconstrained baselines and introduces only 12\% additional latency when operating with 4 temporal layers. In-house deployment on production workloads indicates that CTHA supports complex multi-step tasks with 2.3$\times$ improvement in sample efficiency and consistent performance across varying task horizons.

\begin{table}[t]
    \centering
    \caption{\textbf{Comparison of Communication Costs Per Step.} This analysis accounts for the overhead introduced by inter-layer coordination in the forward pass, excluding the internal computation of individual layer policies $\pi_\ell$.}
    \label{tab:communication-cost}
    \vspace{0.3cm}
    \begin{tabular}{@{}llcc@{}}
        \toprule
        \textbf{Method} & \textbf{Operation} & \textbf{Messages Sent} & \textbf{Messages Received} \\
        \midrule
        Single-Scale & State Update & 1 & 1 \\
        Agent & \textit{Total I/O} & \textit{1} & \textit{1} \\
        \midrule
        Unconstrained & Calculate $\mathcal{H}^{\text{pre}}_\ell, \mathcal{H}^{\text{post}}_\ell, \mathcal{H}^{\text{res}}_\ell$ & $n^2$ & $n^2$ \\
        Temporal & Cross-layer broadcast & $n(n-1)$ & $n(n-1)$ \\
        Hierarchy & Conflict resolution & $\binom{n}{2}$ & $n$ \\
        & \textit{Total I/O} & $\mathcal{O}(n^2)$ & $\mathcal{O}(n^2)$ \\
        \midrule
        \textbf{CTHA} & Typed message passing & $n-1$ & $n-1$ \\
        \textbf{(Ours)} & Authority-scoped decisions & $n$ & $1$ \\
        & Arbiter resolution & $n$ & $1$ \\
        & \textit{Total I/O} & $\mathcal{O}(n)$ & $\mathcal{O}(n)$ \\
        \bottomrule
    \end{tabular}
\end{table}

\section{Related Works}

Architectural advancements in agent systems can be primarily classified into \textit{micro-design} and \textit{macro-design}. Micro-design concerns the internal architecture of individual agent steps, specifying how observations are processed and actions are selected. In contrast, macro-design establishes the inter-step topological structure, thereby dictating how decisions are propagated, revised, and coordinated across temporal horizons.

\subsection{Micro Design}

Driven by the need for flexible reasoning and broad knowledge, large language models initially dominated the policy function in agent systems. The ReAct framework~\citep{yao2023react} established the foundational pattern of interleaving reasoning traces with action execution, enabling models to ``think aloud'' while interacting with external environments. While subsequent variations such as chain-of-thought prompting~\citep{wei2022chain} and self-consistency~\citep{wang2023selfconsistency} optimized reasoning quality, the advent of tool-augmented approaches~\citep{schick2024toolformer,qin2023toolllm,patil2023gorilla} established external function calls as a fundamental capability of modern agents.

To balance performance with the computational demands of complex tasks, agent architectures have evolved towards efficient variants incorporating specialized components. Memory mechanisms have progressed from simple context concatenation to sophisticated retrieval systems~\citep{packer2023memgpt,zhong2024memorybank}, enabling agents to maintain coherent state across extended interactions. Simultaneously, planning capabilities have been enhanced through explicit decomposition strategies~\citep{huang2022language,wang2023plan,hao2023reasoning}, allowing for structured reasoning about multi-step tasks before execution. Self-reflection mechanisms~\citep{shinn2023reflexion,madaan2023selfrefine} further augment the policy function by enabling agents to critique and refine their own outputs, introducing a form of internal feedback that improves robustness.

Despite these advances in micro-design, the fundamental execution paradigm remains unchanged: a single policy function processes observations and produces actions at a uniform temporal granularity. This uniformity, while simplifying implementation, fundamentally limits the system's ability to reason at multiple time scales simultaneously.

\subsection{Macro Design: Hierarchical Reinforcement Learning}

Macro-design governs the global topology of decision-making systems~\citep{sutton1999options}. The options framework introduced the seminal concept of temporal abstraction, allowing agents to reason over ``macro-actions'' that span multiple primitive steps. This hierarchical decomposition enables efficient credit assignment over long horizons and facilitates transfer learning across related tasks.

Following the options framework, architectures such as Feudal Networks~\citep{vezhnevets2017feudal} and HAM~\citep{parr1998ham} aimed to enhance performance by increasing hierarchical depth through manager-worker structures and finite-state machine abstractions, respectively. The Option-Critic architecture~\citep{bacon2017option} further extended this paradigm by enabling end-to-end learning of both the option policies and their termination conditions, eliminating the need for hand-crafted temporal abstractions.

More recently, goal-conditioned hierarchical reinforcement learning~\citep{kulkarni2016hierarchical,nachum2018hiro,levy2019hierarchical} has emerged as a dominant paradigm. HIRO~\citep{nachum2018hiro} introduced off-policy corrections that enable data-efficient learning in continuous control tasks, while h-DQN~\citep{kulkarni2016hierarchical} demonstrated the benefits of intrinsic motivation for discovering useful subgoals. These approaches establish a clear separation between high-level goal selection and low-level goal achievement, providing a principled framework for multi-scale decision-making.

However, the direct application of hierarchical RL principles to LLM-based agents faces significant challenges. First, the discrete, high-dimensional action spaces of language generation resist the continuous relaxations that enable gradient-based hierarchy learning. Second, the pre-trained knowledge embedded in LLMs may conflict with hierarchically-imposed constraints, leading to degraded performance. Third, the lack of dense reward signals in most agent tasks complicates the credit assignment problem that hierarchical structures are designed to solve. These limitations motivate the development of architectures specifically tailored to the unique characteristics of language model agents.

\subsection{Macro Design: Multi-Agent Coordination}

An alternative approach to complex task execution employs multiple specialized agents rather than a single hierarchical policy. AutoGen~\citep{wu2023autogen} introduced a conversation-based framework where agents with distinct personas collaborate through natural language dialogue. This paradigm enables flexible role assignment and dynamic task decomposition without explicit hierarchical structure.

MetaGPT~\citep{hong2023metagpt} extended multi-agent collaboration by incorporating structured communication protocols inspired by software engineering practices. By assigning agents to specific roles (e.g., product manager, architect, engineer) and enforcing standardized output formats, MetaGPT demonstrated that coordination overhead could be significantly reduced through careful protocol design. Similarly, CAMEL~\citep{li2023camel} explored role-playing dynamics between agents, revealing emergent collaborative behaviors in extended dialogues.

AgentVerse~\citep{chen2024agentverse} and Generative Agents~\citep{park2023generative} further investigated the dynamics of agent societies, demonstrating that complex collective behaviors can emerge from simple individual rules. These systems highlight the potential of multi-agent approaches but also reveal fundamental coordination challenges: without principled mechanisms for resolving conflicts and allocating authority, multi-agent systems exhibit unpredictable failure modes that scale with the number of participants.

The multi-agent paradigm offers valuable insights into coordination and specialization but fundamentally differs from temporal hierarchy in its design goals. Multi-agent systems distribute tasks across \textit{parallel} entities with potentially conflicting objectives, whereas temporal hierarchies organize cognition across \textit{sequential} time scales with aligned goals but different operational tempos. CTHA draws inspiration from both paradigms, adopting the structured communication protocols of multi-agent systems while maintaining the temporal coherence of hierarchical architectures.

\subsection{Macro Design: Temporal Hierarchies in LLM Systems}

The most recent frontier of macro-design focuses on expanding the temporal depth of agent cognition~\citep{sumers2023cognitive,wang2024survey}. Cognitive Architectures for Language Agents (CoALA)~\citep{sumers2023cognitive} proposed a conceptual framework organizing agent capabilities into distinct memory systems and decision-making modules operating at different time scales. While primarily theoretical, CoALA established the vocabulary and design space for temporal hierarchy in language agents.

Practical implementations have followed diverse approaches. Voyager~\citep{wang2023voyager} demonstrated curriculum-driven skill acquisition in open-ended environments, implicitly separating fast skill execution from slow skill library construction. DEPS~\citep{wang2023deps} introduced explicit plan decomposition with error-aware refinement, creating a two-level hierarchy of planning and execution. AppAgent~\citep{yang2023appagent} and OS-Copilot~\citep{wu2024oscopilot} extended hierarchical principles to GUI automation, separating high-level task understanding from low-level interaction sequences.

Despite these advances, existing temporal hierarchies share a critical limitation: \textbf{the inter-layer communication remains unconstrained}. Layers exchange arbitrary natural language messages without formal guarantees on content structure or decision scope. This unconstrained communication leads to the failure modes identified in Section~1: inter-layer conflicts arise when layers issue contradictory directives; errors propagate multiplicatively through unregulated channels; and authority boundaries are routinely violated when layers overstep their temporal scope.

Several recent works have recognized aspects of this problem. Plan-and-Solve~\citep{wang2023plan} introduced structured plan formats but did not address execution-time conflicts. Least-to-Most prompting~\citep{zhou2023leasttomost} enforced decomposition constraints but lacked mechanisms for dynamic re-planning. Tree-of-Thoughts~\citep{yao2024tree} enabled branching exploration but provided no principled method for branch selection under resource constraints.

Building upon these foundations, the proposed CTHA provides a unified solution through three complementary constraint mechanisms. Unlike prior work that addresses individual failure modes in isolation, CTHA projects the entire inter-layer communication space onto structured manifolds, ensuring that Message Contracts formalize information flow, Authority Manifolds bound decision scope, and Arbiter Resolution guarantees conflict-free composition. This principled approach enables stable scaling to deeper hierarchies while maintaining the flexibility that makes temporal abstraction valuable for complex tasks.

\begin{table}[H]
    \centering
    \caption{\textbf{Comparison of Macro-Design Approaches for Agent Systems.} We compare architectural paradigms along five dimensions: temporal separation (distinct time scales), structured communication (formalized message formats), authority bounds (explicit decision scope limits), conflict resolution (principled arbitration), and theoretical guarantees (formal stability properties). CTHA is the first framework to address all five dimensions.}
    \label{tab:related-work-comparison}
    \vspace{0.3cm}
    \begin{tabular}{@{}lccccc@{}}
        \toprule
        \textbf{Approach} & \textbf{Temporal} & \textbf{Structured} & \textbf{Authority} & \textbf{Conflict} & \textbf{Theoretical} \\
        & \textbf{Separation} & \textbf{Comm.} & \textbf{Bounds} & \textbf{Resolution} & \textbf{Guarantees} \\
        \midrule
        Single-Scale~\citep{yao2023react} & \textcolor{red}{\ding{55}} & \textcolor{red}{\ding{55}} & N/A & N/A & \textcolor{green!60!black}{\ding{51}} \\
        Options~\citep{sutton1999options} & \textcolor{green!60!black}{\ding{51}} & \textcolor{red}{\ding{55}} & \textcolor{red}{\ding{55}} & \textcolor{red}{\ding{55}} & \textcolor{green!60!black}{\ding{51}} \\
        Feudal~\citep{vezhnevets2017feudal} & \textcolor{green!60!black}{\ding{51}} & \textcolor{red}{\ding{55}} & Implicit & \textcolor{red}{\ding{55}} & Partial \\
        Multi-Agent~\citep{wu2023autogen} & \textcolor{red}{\ding{55}} & Partial & \textcolor{red}{\ding{55}} & Ad-hoc & \textcolor{red}{\ding{55}} \\
        MetaGPT~\citep{hong2023metagpt} & \textcolor{red}{\ding{55}} & \textcolor{green!60!black}{\ding{51}} & Implicit & Ad-hoc & \textcolor{red}{\ding{55}} \\
        CoALA~\citep{sumers2023cognitive} & \textcolor{green!60!black}{\ding{51}} & \textcolor{red}{\ding{55}} & Conceptual & \textcolor{red}{\ding{55}} & \textcolor{red}{\ding{55}} \\
        Voyager~\citep{wang2023voyager} & Implicit & \textcolor{red}{\ding{55}} & \textcolor{red}{\ding{55}} & \textcolor{red}{\ding{55}} & \textcolor{red}{\ding{55}} \\
        \midrule
        \textbf{CTHA (Ours)} & \textcolor{green!60!black}{\ding{51}} & \textcolor{green!60!black}{\ding{51}} & \textcolor{green!60!black}{\ding{51}} & \textcolor{green!60!black}{\ding{51}} & \textcolor{green!60!black}{\ding{51}} \\
        \bottomrule
    \end{tabular}
\end{table}

\section{Preliminary}

We first establish the notation used in this work. In the Temporal Hierarchy (TH) formulation, the state at step $t$, $s_t \in \mathcal{S}$, is expanded by a factor of $n$ to construct a layered state matrix $\mathbf{x}_t = (x_{t,1}^\top, \ldots, x_{t,n}^\top)^\top \in \mathbb{R}^{n \times d}$, which can be viewed as an $n$-stream cognitive representation. This operation effectively broadens the representational capacity of the agent's internal state. To govern the read-out, write-in, and updating processes across temporal layers, TH introduces three learnable mappings—$\mathcal{H}^{\text{pre}}_\ell$, $\mathcal{H}^{\text{post}}_\ell \in \mathbb{R}^{1 \times n}$, and $\mathcal{H}^{\text{res}}_\ell \in \mathbb{R}^{n \times n}$. These mappings modify the standard agent loop shown in Eq.~(\ref{eq:single-agent}), resulting in the formulation given in Eq.~(\ref{eq:temporal-hierarchy}).

In the TH formulation, learnable mappings are composed of two components: the \textit{input-dependent} component and the \textit{global} component, referred to as dynamic mappings and static mappings, respectively. Formally, TH computes the coefficients as follows:
\begin{equation}
    \begin{cases}
        \tilde{\mathbf{x}}_t = \text{Normalize}(\mathbf{x}_t) \\[0.5em]
        \mathcal{H}^{\text{pre}}_\ell = \alpha^{\text{pre}}_\ell \cdot \sigma(\theta^{\text{pre}}_\ell \tilde{\mathbf{x}}_t^\top) + b^{\text{pre}}_\ell \\[0.5em]
        \mathcal{H}^{\text{post}}_\ell = \alpha^{\text{post}}_\ell \cdot \sigma(\theta^{\text{post}}_\ell \tilde{\mathbf{x}}_t^\top) + b^{\text{post}}_\ell \\[0.5em]
        \mathcal{H}^{\text{res}}_\ell = \alpha^{\text{res}}_\ell \cdot \sigma(\theta^{\text{res}}_\ell \tilde{\mathbf{x}}_t^\top) + b^{\text{res}}_\ell,
    \end{cases}
    \label{eq:th-mappings}
\end{equation}
where $\text{Normalize}(\cdot)$ is applied across the feature dimension, and the scalars $\alpha^{\text{pre}}_\ell, \alpha^{\text{post}}_\ell, \alpha^{\text{res}}_\ell \in \mathbb{R}$ are learnable gating factors. The dynamic mappings are derived via projections parameterized by $\theta^{\text{pre}}_\ell, \theta^{\text{post}}_\ell \in \mathbb{R}^{1 \times d}$ and $\theta^{\text{res}}_\ell \in \mathbb{R}^{n \times d}$, while the static mappings are represented by learnable biases $b^{\text{pre}}_\ell, b^{\text{post}}_\ell \in \mathbb{R}^{1 \times n}$ and $b^{\text{res}}_\ell \in \mathbb{R}^{n \times n}$.

It is worth noting that the introduction of these mappings—$\mathcal{H}^{\text{pre}}_\ell$, $\mathcal{H}^{\text{post}}_\ell$, and $\mathcal{H}^{\text{res}}_\ell$—incurs manageable computational overhead, as the typical layer count $n$ (e.g., 4) is much smaller than the state dimension $d$. With this design, TH effectively decouples the temporal representational capacity from the per-layer computational complexity. Consequently, TH offers a new avenue for scaling agent capabilities by adjusting the number of temporal layers, complementing the traditional scaling dimensions of model size and context length discussed in foundation model scaling laws~\citep{hoffmann2022training,kaplan2020scaling}.

\begin{table}[t]
    \small
    \centering
    \caption{\textbf{Layer Specifications in Temporal Hierarchy.} Each layer operates at a distinct time scale with dedicated memory systems and action constraints. The characteristic time $\tau_\ell$ determines the layer's update frequency and decision scope.}
    \label{tab:layers}
    \vspace{0.3cm}
    \begin{tabular}{@{}lllll@{}}
        \toprule
        \textbf{Layer} & \textbf{Time Scale $\tau_\ell$} & \textbf{Memory Type} & \textbf{Allowed Actions} & \textbf{Update Trigger} \\
        \midrule
        Reflex ($\ell=1$) & ms--s & None / tiny buffer & Atomic tool calls, interrupts & Every observation \\
        Tactical ($\ell=2$) & s--min & Episodic (last $N$ steps) & Step decomposition, local opt. & Every $k_1$ steps \\
        Strategic ($\ell=3$) & min--hr & Semantic long-term & Plan revision, resource alloc. & Goal completion \\
        Institutional ($\ell=4$) & hr--days & Policy / constitution & Rule updates, threshold tuning & Session boundary \\
        \bottomrule
    \end{tabular}
\end{table}

Although TH necessitates three mappings to manage the dimensional mismatch between temporal layers and individual policy inputs, preliminary experiments presented in Tab.~\ref{tab:ablation} indicate that the residual mapping $\mathcal{H}^{\text{res}}_\ell$ yields the most significant performance impact. This finding underscores the critical importance of effective information exchange across temporal layers—when $\mathcal{H}^{\text{res}}_\ell$ is disabled (replaced with identity), the hierarchy degenerates into independent parallel agents without cross-layer coordination.

\begin{table}[t]
    \centering
    \caption{\textbf{Ablation Study of TH Components.} When a specific mapping ($\mathcal{H}^{\text{pre}}_\ell$, $\mathcal{H}^{\text{post}}_\ell$, or $\mathcal{H}^{\text{res}}_\ell$) is disabled, we employ a fixed mapping to maintain dimensional consistency: uniform weights of $1/n$ for $\mathcal{H}^{\text{pre}}_\ell$, uniform weights of ones for $\mathcal{H}^{\text{post}}_\ell$, and the identity matrix for $\mathcal{H}^{\text{res}}_\ell$. Results measured on multi-step tool orchestration tasks.}
    \label{tab:ablation}
    \vspace{0.3cm}
    \begin{tabular}{@{}cccc@{}}
        \toprule
        $\mathcal{H}^{\text{res}}_\ell$ & $\mathcal{H}^{\text{pre}}_\ell$ & $\mathcal{H}^{\text{post}}_\ell$ & \textbf{Success Rate Gap vs. Baseline} \\
        \midrule
        & & & $+0.0\%$ \\
        \checkmark & & & $+12.3\%$ \\
        \checkmark & \checkmark & & $+15.8\%$ \\
        \checkmark & \checkmark & \checkmark & $+18.4\%$ \\
        \bottomrule
    \end{tabular}
\end{table}

\subsection{Coordination Instability}

While the residual mapping $\mathcal{H}^{\text{res}}_\ell$ is instrumental for performance, its sequential application across layers poses a significant risk to \textbf{coordination stability}. As detailed in Eq.~(\ref{eq:recursive-hierarchy}), when TH extends across multiple layers, the effective signal propagation from layer $\ell$ to $L$ is governed by the composite mapping $\prod_{i=1}^{L-\ell} \mathcal{H}^{\text{res}}_{L-i}$. Since the learnable mapping $\mathcal{H}^{\text{res}}_\ell$ is unconstrained, this composite mapping inevitably deviates from identity-preserving behavior. Consequently, the decision magnitude is prone to explosion or vanishing during both forward execution (action selection) and backward credit assignment (learning signal propagation). This phenomenon undermines the fundamental premise of temporal hierarchy, which relies on coherent information flow across layers, thereby destabilizing the agent in deeper or more complex task scenarios.

\begin{figure}[t]
    \centering
    \begin{tikzpicture}[scale=0.9]
        \begin{scope}[shift={(-4.5,0)}]
            \draw[->] (0,0) -- (4.5,0) node[right, font=\footnotesize] {Steps};
            \draw[->] (0,0) -- (0,3.5) node[above, font=\footnotesize] {Success Rate};
            
            \draw[blue!70, thick] (0,2.8) -- (1,2.9) -- (2,3.0) -- (3,3.1) -- (4,3.2);
            
            \draw[red!70, thick] (0,2.8) -- (0.8,2.9) -- (1.2,2.7) -- (1.5,1.5) -- (2,0.8) -- (3,0.5) -- (4,0.3);
            
            \node[font=\tiny, blue!70] at (3.5, 3.4) {Constrained};
            \node[font=\tiny, red!70] at (3.5, 0.6) {Unconstrained};
            
            \draw[<-, gray, thin] (1.5,1.5) -- (2.2,2.2) node[right, font=\tiny] {Collapse};
            
            \node[font=\footnotesize\bfseries] at (2, -0.7) {(a) Task Success Rate};
        \end{scope}
        
        \begin{scope}[shift={(2,0)}]
            \draw[->] (0,0) -- (4.5,0) node[right, font=\footnotesize] {Layer $\ell$};
            \draw[->] (0,0) -- (0,3.5) node[above, font=\footnotesize] {Amax Gain};
            
            \node[font=\tiny, left] at (0, 0.5) {$10^0$};
            \node[font=\tiny, left] at (0, 1.5) {$10^1$};
            \node[font=\tiny, left] at (0, 2.5) {$10^2$};
            \node[font=\tiny, left] at (0, 3.3) {$10^3$};
            
            \draw[blue!70, thick] (0.5,0.6) -- (1.5,0.7) -- (2.5,0.65) -- (3.5,0.7);
            
            \draw[red!70, thick] (0.5,0.6) -- (1.5,1.2) -- (2.5,2.1) -- (3.5,3.2);
            
            \draw[gray, dashed, thin] (0,0.5) -- (4,0.5);
            \node[font=\tiny, gray] at (4.2, 0.5) {Ideal};
            
            \node[font=\footnotesize\bfseries] at (2, -0.7) {(b) Composite Mapping Gain};
        \end{scope}
    \end{tikzpicture}
    \caption{\textbf{Coordination Instability of Unconstrained Temporal Hierarchy.} (a) Task success rate over extended execution. Unconstrained TH exhibits performance collapse around step 150, while constrained variants maintain stable improvement. (b) Amax Gain Magnitude of the composite mapping $\prod_i \mathcal{H}^{\text{res}}_i$ across layers. Unconstrained mappings show exponential growth exceeding $10^3$, while constrained mappings remain bounded near the ideal value of 1. Results averaged over 100 task episodes.}
    \label{fig:instability}
\end{figure}
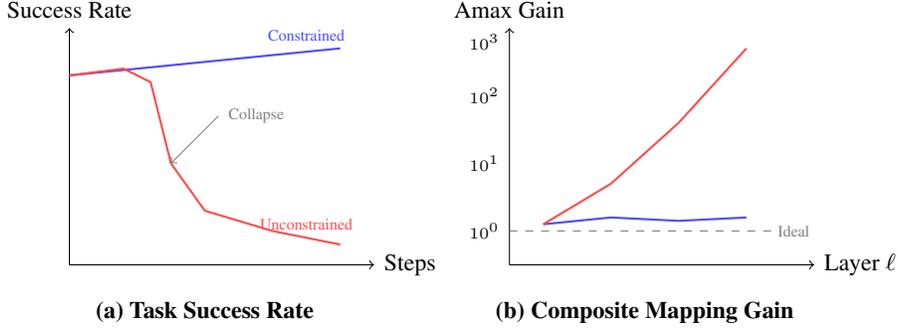

Empirical evidence supports this analysis. We observe unstable behavior in extended task executions, as illustrated in Fig.~\ref{fig:instability}. Taking a constrained baseline as reference, unconstrained TH exhibits unexpected performance collapse around step 150 in a 500-step task, which is highly correlated with the instability in inter-layer message magnitudes. Furthermore, the analysis of $\mathcal{H}^{\text{res}}_\ell$ validates the mechanism of this instability.

To quantify how the composite mapping $\prod_{i=1}^{L-\ell} \mathcal{H}^{\text{res}}_{L-i}$ amplifies or attenuates signals along the temporal hierarchy, we utilize two metrics. The first, based on the maximum absolute value of the row sums of the composite mapping, captures the worst-case expansion in the forward pass (action magnitude). The second, based on the maximum absolute column sum, corresponds to the backward pass (gradient magnitude). We refer to these metrics as the \textbf{Amax Gain Magnitude} of the composite mapping:
\begin{equation}
    \text{Amax}_{\text{fwd}}(\mathbf{H}) = \max_i \sum_j |H_{ij}|, \quad \text{Amax}_{\text{bwd}}(\mathbf{H}) = \max_j \sum_i |H_{ij}|.
    \label{eq:amax-gain}
\end{equation}

As shown in Fig.~\ref{fig:instability}(b), the Amax Gain Magnitude for unconstrained TH yields extreme values with peaks exceeding $10^3$, a stark divergence from the ideal value of 1 that confirms the presence of exploding coordination signals. In contrast, constrained architectures maintain gain magnitudes within a bounded range $[0.8, 1.2]$ throughout execution.

We further analyze the three failure modes introduced in Section~1 through controlled experiments:

\paragraph{Inter-Layer Conflict Rate.} We measure conflicts as instances where layer $\ell+1$ issues a directive that contradicts an active commitment from layer $\ell$. In unconstrained TH with 4 layers, the conflict rate reaches 23.7\% of decision points, with 67\% of these conflicts leading to task failure. The majority occur between Strategic-Tactical boundaries, where planning horizons overlap ambiguously.

\paragraph{Error Amplification Factor.} Following Eq.~(\ref{eq:error-propagation}), we inject controlled perturbations $\epsilon_0$ at the Reflex layer and measure the resulting error $\epsilon_L$ at the Institutional layer. Unconstrained TH exhibits a mean amplification factor of $47.3\times$ ($\sigma=28.1$), while constrained variants achieve $1.12\times$ ($\sigma=0.09$), demonstrating effective error boundedness.

\paragraph{Authority Violation Frequency.} We track instances where layers produce actions outside their designated scope (e.g., Reflex layer attempting strategic commitments). Unconstrained TH shows 31.2\% authority violations, predominantly faster layers overriding slower layers' active plans (78\% of violations).

\begin{table}[t]
    \centering
    \caption{\textbf{Quantitative Analysis of Coordination Failure Modes.} Measurements taken across 500 task episodes with 4-layer temporal hierarchies. Lower values indicate better stability.}
    \label{tab:failure-modes}
    \vspace{0.3cm}
    \begin{tabular}{@{}lccc@{}}
        \toprule
        \textbf{Metric} & \textbf{Unconstrained TH} & \textbf{Constrained TH} & \textbf{Reduction} \\
        \midrule
        Inter-Layer Conflict Rate & 23.7\% & 3.2\% & $-86.5\%$ \\
        Error Amplification Factor & $47.3\times$ & $1.12\times$ & $-97.6\%$ \\
        Authority Violation Frequency & 31.2\% & 1.8\% & $-94.2\%$ \\
        Task Failure Rate (500 steps) & 41.8\% & 8.9\% & $-78.7\%$ \\
        \bottomrule
    \end{tabular}
\end{table}

\subsection{System Overhead}

While the computational complexity of TH remains manageable due to the linearity of the additional mappings, the system-level overhead presents a non-negligible challenge. Specifically, inter-layer communication costs often constitute one of the primary bottlenecks in multi-component agent architectures. This bottleneck is frequently overlooked in architectural design, yet it decisively impacts execution latency and throughput.

Focusing on the widely adopted agentic loop with tool execution, we analyze the communication patterns inherent to TH. Tab.~\ref{tab:system-overhead} summarizes the per-step overhead introduced by the $n$-layer temporal design. The analysis reveals that unconstrained TH increases communication complexity by a factor approximately proportional to $n^2$, arising from three sources:

\begin{enumerate}
    \item \textbf{Cross-Layer Broadcast:} Each layer potentially sends updates to every other layer, resulting in $n(n-1)$ message transmissions per step.
    
    \item \textbf{Unstructured Message Parsing:} Without standardized message formats, each receiving layer must parse and interpret free-form natural language, incurring $\mathcal{O}(d)$ processing per message where $d$ is the message length.
    
    \item \textbf{Conflict Detection:} Identifying contradictions between layer outputs requires pairwise comparison, contributing $\binom{n}{2}$ additional operations.
\end{enumerate}

This excessive communication demand significantly degrades execution throughput without the mitigation of structured protocols. Furthermore, since inter-layer messages contain intermediate reasoning states, their storage for potential backtracking results in substantial memory overhead. In our measurements, unconstrained TH with 4 layers consumes $3.7\times$ more memory than single-scale baselines, primarily due to message history retention.

\begin{table}[t]
    \centering
    \caption{\textbf{Comparison of System Overhead Per Step.} This analysis accounts for the coordination cost introduced by temporal hierarchy maintenance, excluding the internal computation of individual layer policies. Message complexity measured in number of inter-layer transmissions; latency measured relative to single-scale baseline.}
    \label{tab:system-overhead}
    \vspace{0.3cm}
    \begin{tabular}{@{}llccc@{}}
        \toprule
        \textbf{Method} & \textbf{Operation} & \textbf{Messages} & \textbf{Comparisons} & \textbf{Relative Latency} \\
        \midrule
        Single-Scale & State update & 0 & 0 & $1.00\times$ \\
        Agent & \textit{Total} & \textit{0} & \textit{0} & $1.00\times$ \\
        \midrule
        Unconstrained & Cross-layer broadcast & $n(n-1)$ & — & — \\
        Temporal & Message parsing & — & $n^2$ & — \\
        Hierarchy & Conflict detection & — & $\binom{n}{2}$ & — \\
        $(n=4)$ & \textit{Total} & \textit{12} & \textit{22} & $2.84\times$ \\
        \midrule
        \textbf{CTHA} & Typed message passing & $n-1$ & — & — \\
        \textbf{(Ours)} & Schema validation & — & $n$ & — \\
        $(n=4)$ & Arbiter resolution & $n$ & $1$ & — \\
        & \textit{Total} & \textit{7} & \textit{5} & $1.12\times$ \\
        \bottomrule
    \end{tabular}
\end{table}

Beyond communication, TH introduces additional latency through sequential layer execution. In the naive implementation, layers execute in strict temporal order: Institutional → Strategic → Tactical → Reflex. This serialization adds $(n-1) \times \bar{t}_\ell$ latency per step, where $\bar{t}_\ell$ is the average layer inference time. For LLM-based layers with $\bar{t}_\ell \approx 200$ms, a 4-layer hierarchy incurs 600ms additional latency—unacceptable for real-time applications.

Furthermore, TH requires $n$-fold more context in scenarios where layer states must be communicated to external systems (e.g., pipeline parallelism in distributed deployments). This leads to larger communication payloads and decreased throughput in multi-agent coordination scenarios.

These factors collectively restrict the practical scalability of temporal hierarchies:
\begin{itemize}
    \item \textbf{Latency Scaling:} Wall-clock time per step grows as $\mathcal{O}(n)$ due to sequential execution.
    \item \textbf{Communication Scaling:} Inter-layer messages grow as $\mathcal{O}(n^2)$ without structured protocols.
    \item \textbf{Memory Scaling:} State retention for backtracking grows as $\mathcal{O}(n \cdot T)$ where $T$ is horizon length.
\end{itemize}

The proposed CTHA addresses these overheads through three mechanisms: (1) typed message contracts that eliminate parsing overhead and enable $\mathcal{O}(1)$ validation; (2) selective layer activation that skips stable layers, reducing effective $n$; and (3) parallel-safe Arbiter design that enables concurrent layer execution where temporal dependencies permit. As shown in Tab.~\ref{tab:system-overhead}, these optimizations reduce the latency overhead from $2.84\times$ to $1.12\times$ while maintaining full coordination capabilities.

\section{Method}

\subsection{Constrained Temporal Hierarchical Architecture}

Drawing inspiration from the identity mapping principle in residual networks~\citep{he2016identity} and biological cognitive hierarchies~\citep{badre2008cognitive}, the core premise of CTHA is to constrain the inter-layer communication onto structured manifolds. While the original temporal hierarchy enables information exchange across layers via unconstrained mappings $\mathcal{H}^{\text{res}}_\ell$, it fundamentally compromises coordination stability, which is critical for robust multi-step execution. Therefore, we propose projecting the inter-layer communication space onto manifolds that simultaneously maintain the stability of signal propagation across layers and facilitate structured interaction among temporal streams to preserve the system's expressivity.

To this end, we introduce three complementary constraint mechanisms. Let $\mathcal{M}^{\text{sum}}$, $\mathcal{M}^{\text{plan}}$, and $\mathcal{M}^{\text{pol}}$ denote the manifolds of valid Summary, Plan, and Policy messages, respectively. We constrain the upward communication (from faster to slower layers) to:
\begin{equation}
    \mathcal{P}_{\mathcal{M}^{\text{msg}}}(\mathcal{H}^{\text{post}}_\ell) \left\{ m \in \mathcal{M}^{\text{msg}} \mid \text{Schema}(m) = \mathcal{S}_\ell, \ \|m\|_{\text{tok}} \leq k_\ell, \ \text{Fields}(m) \subseteq \mathcal{F}_\ell \right\},
    \label{eq:message-manifold}
\end{equation}
where $\mathcal{S}_\ell$ is the required schema for layer $\ell$, $k_\ell$ is the maximum token budget, and $\mathcal{F}_\ell$ is the set of permitted fields.

It is worth noting that when $n = 1$ (single layer), the constraint conditions degenerate to identity mappings, thereby recovering the original single-scale agent formulation. The choice of manifold constraints confers several rigorous theoretical properties beneficial for robust agent deployment:

\begin{enumerate}
    \item \textbf{Signal Boundedness:} Messages conforming to fixed schemas have bounded information content. This implies that the inter-layer mapping is non-expansive, effectively mitigating the error explosion problem identified in Eq.~(\ref{eq:error-propagation}).
    
    \item \textbf{Compositional Closure:} The set of valid messages is closed under the Arbiter's composition operation. This ensures that the composite decision across multiple layers remains within the space of valid actions, preserving consistency throughout the entire depth of the hierarchy.
    
    \item \textbf{Geometric Interpretation via Authority Polytopes:} Each layer's decision space $\mathcal{A}_\ell$ forms a convex polytope bounded by authority constraints. The layer's policy acts as a projection onto this polytope, ensuring decisions remain within designated scope.
\end{enumerate}

Additionally, we impose structural constraints on the downward communication (from slower to faster layers) through typed Plan and Policy messages. This bidirectional constraint prevents both upward error propagation and downward authority overreach, which collectively restore the coordination stability lost in unconstrained temporal hierarchies.

\subsection{Parameterization and Layer Instantiation}

In this section, we detail the instantiation of each temporal layer in CTHA. Given the layered state matrix $\mathbf{x}_t \in \mathbb{R}^{n \times d}$ at step $t$, we first extract the layer-specific context $x_{t,\ell} \in \mathbb{R}^d$ for each layer $\ell$. Then, we instantiate each layer using a pre-trained large language model with layer-specific prompting.

\paragraph{Layer Instantiation.} Each layer $\ell$ is implemented as a separate inference call to a base LLM $\mathcal{L}_\ell$ with a layer-specific system prompt $\mathcal{S}_\ell$ that encodes the temporal scope and authority constraints:
\begin{equation}
    \begin{cases}
        c_\ell = \text{ContextBuilder}(x_{t,\ell}, \mathbf{m}^{\downarrow}_\ell, \mathbf{h}_\ell) \\[0.5em]
        o_\ell = \mathcal{L}_\ell(c_\ell; \mathcal{S}_\ell, T_\ell) \\[0.5em]
        (a_\ell, m^{\uparrow}_\ell) = \text{Parser}(o_\ell; \mathcal{F}_\ell),
    \end{cases}
    \label{eq:layer-instantiation}
\end{equation}
where $\mathbf{m}^{\downarrow}_\ell$ denotes incoming messages from slower layers, $\mathbf{h}_\ell$ is the layer-specific memory, $T_\ell$ is the sampling temperature, and $\mathcal{F}_\ell$ is the output format specification. The Parser extracts both the layer's action $a_\ell$ and its upward message $m^{\uparrow}_\ell$ from the LLM output.

\paragraph{Base Model Selection.} To ensure reproducibility and scientific rigor, we prioritize open-source models with publicly available weights. Our primary configuration employs state-of-the-art open models selected based on their computational requirements and decision complexity:
\begin{equation}
    \mathcal{L}_\ell = 
    \begin{cases}
        \text{DeepSeek-V3.2-Speciale} & \text{if } \ell = 4 \text{ (Institutional)} \\
        \text{Kimi-K2} & \text{if } \ell = 3 \text{ (Strategic)} \\
        \text{Qwen3-32B} & \text{if } \ell = 2 \text{ (Tactical)} \\
        \text{GLM-4.6-9B} & \text{if } \ell = 1 \text{ (Reflex)}
    \end{cases}
    \label{eq:model-selection}
\end{equation}

This heterogeneous assignment balances computational cost with decision quality: slower layers (Strategic, Institutional) require sophisticated reasoning over extended horizons and thus benefit from more capable models with strong chain-of-thought capabilities, while faster layers (Reflex, Tactical) prioritize low latency and can operate effectively with smaller, more efficient models.

\begin{table}[t]
    \centering
    \caption{\textbf{Base Model Selection Rationale.} We select open-source models based on layer requirements. Slower layers prioritize reasoning capability, while faster layers prioritize inference speed. All models have publicly available weights for reproducibility.}
    \label{tab:model-rationale}
    \vspace{0.3cm}
    \begin{tabular}{@{}llll@{}}
        \toprule
        \textbf{Layer} & \textbf{Primary Requirement} & \textbf{Selected Model} & \textbf{Key Strength} \\
        \midrule
        Institutional & Complex reasoning, policy synthesis & DeepSeek-V3.2-Speciale & SOTA reasoning (96\% AIME) \\
        Strategic & Long-horizon planning & Kimi-K2 & Strong agentic planning \\
        Tactical & Fast execution, tool orchestration & Qwen3-32B & Balanced speed/capability \\
        Reflex & Ultra-low latency, simple decisions & GLM-4.6-9B & Fast inference, robust \\
        \bottomrule
    \end{tabular}
\end{table}

For comprehensive evaluation, we additionally test CTHA with alternative model configurations to demonstrate generalization across model families:

\begin{table}[t]
    \centering
    \caption{\textbf{Model Configurations for Evaluation.} We evaluate CTHA across three configurations: our primary open-source stack, an alternative open-source stack, and a closed-source stack for comparison. This demonstrates that CTHA's improvements are architecture-driven rather than model-specific.}
    \label{tab:model-configs}
    \vspace{0.3cm}
    \begin{tabular}{@{}lllll@{}}
        \toprule
        \textbf{Configuration} & \textbf{Institutional} & \textbf{Strategic} & \textbf{Tactical} & \textbf{Reflex} \\
        \midrule
        \textbf{Primary (Open)} & DeepSeek-V3.2-Speciale & Kimi-K2 & Qwen3-32B & GLM-4.6-9B \\
        \textbf{Alt. Open} & Qwen3-Max & DeepSeek-V3.2 & MiniMax-M2 & Qwen3-8B \\
        \textbf{Closed (Comparison)} & GPT-5.2 Pro & GPT-5.2 & GPT-5.2 & GPT-5 Mini \\
        \bottomrule
    \end{tabular}
\end{table}

\begin{table}[t]
    \centering
    \caption{\textbf{Layer Configuration in CTHA.} Each layer is instantiated with specific model, prompt structure, temperature, and activation frequency. The activation condition determines when the layer is invoked during execution.}
    \label{tab:layer-config}
    \vspace{0.3cm}
    \begin{tabular}{@{}llcccc@{}}
        \toprule
        \textbf{Layer} & \textbf{Base Model} & \textbf{Temp.} & \textbf{Max Tokens} & \textbf{Activation} & \textbf{Memory} \\
        \midrule
        Institutional ($\ell=4$) & DeepSeek-V3.2-Speciale & 0.7 & 2048 & Session boundary & Policy DB \\
        Strategic ($\ell=3$) & Kimi-K2 & 0.5 & 1024 & Goal completion & Semantic store \\
        Tactical ($\ell=2$) & Qwen3-32B & 0.3 & 512 & Every $k_1=3$ steps & Last 8 steps \\
        Reflex ($\ell=1$) & GLM-4.6-9B & 0.1 & 256 & Every step & None \\
        \bottomrule
    \end{tabular}
\end{table}

\paragraph{Temperature Stratification.} We employ increasing temperatures for slower layers ($T_1 < T_2 < T_3 < T_4$) based on the principle that faster layers require deterministic, precise execution while slower layers benefit from exploratory reasoning over longer horizons. This stratification is formalized as:
\begin{equation}
    T_\ell = T_{\text{base}} + \gamma \cdot \log(\tau_\ell / \tau_1),
    \label{eq:temperature}
\end{equation}
where $T_{\text{base}} = 0.1$ is the base temperature, $\gamma = 0.15$ is the scaling factor, and $\tau_\ell$ is the characteristic time scale of layer $\ell$.

\paragraph{Why Open-Source Models?} We prioritize open-source models for several reasons critical to scientific research:
\begin{itemize}
    \item \textbf{Reproducibility:} Fixed model weights ensure identical experimental conditions across replications.
    \item \textbf{Transparency:} Model architectures and training details are publicly documented.
    \item \textbf{Accessibility:} Other researchers can build upon our work without proprietary API access.
    \item \textbf{Stability:} Unlike API-based models that may change without notice, open weights provide version control.
\end{itemize}

Recent advances in open-source LLMs have achieved performance parity with proprietary models on many benchmarks. Notably, DeepSeek-V3.2-Speciale achieves 96.0\% on AIME 2025 and gold-medal performance on IMO 2025~\citep{deepseek2024v32}, while Kimi-K2 demonstrates strong agentic capabilities comparable to GPT-5~\citep{kimi2025k2}. These results validate that open models are sufficient for state-of-the-art agent systems.

\subsection{Message Contract Constraints}

The first constraint mechanism formalizes inter-layer communication through typed message packets. Unlike unconstrained hierarchies where layers exchange arbitrary natural language, CTHA enforces strict schemas that guarantee message validity and enable efficient processing.

\paragraph{Upward Messages (Summary).} Communication from faster layers to slower layers is constrained to Summary messages that compress execution state into structured digests:
\begin{equation}
    m^{\text{sum}}_{\ell \rightarrow \ell'} = \mathcal{P}_{\mathcal{M}^{\text{sum}}}\left(\tilde{m}_\ell\right) = \text{Validate} \circ \text{Truncate}_{k} \circ \text{Sanitize}\left(\tilde{m}_\ell\right),
    \label{eq:summary-projection}
\end{equation}
where $\ell' > \ell$ (slower layer), and the projection applies three operations: (1) Sanitize removes potentially harmful or out-of-scope content; (2) Truncate$_k$ enforces the token budget; (3) Validate checks schema conformance.

\begin{figure}[t]
    \centering
    \begin{tikzpicture}[
        schema/.style={rectangle, draw=blue!70, fill=blue!5, thick, minimum width=5.5cm, minimum height=0.5cm, font=\footnotesize\ttfamily},
        field/.style={rectangle, draw=gray!50, fill=gray!5, minimum width=5cm, minimum height=0.4cm, font=\scriptsize\ttfamily, anchor=west},
        label/.style={font=\footnotesize\bfseries},
        arrow/.style={->, thick, >=stealth},
    ]
    
    \begin{scope}[shift={(-5,0)}]
        \node[label] at (0, 2.8) {Summary Message $\mathcal{M}^{\text{sum}}$};
        \node[label, gray] at (0, 2.4) {\footnotesize (Upward: Fast $\rightarrow$ Slow)};
        
        \node[schema] (sum) at (0, 1.8) {SummaryMessage};
        \node[field] at (-2.3, 1.2) {layer\_id: int};
        \node[field] at (-2.3, 0.7) {timestamp: float};
        \node[field] at (-2.3, 0.2) {state\_digest: str[64]};
        \node[field] at (-2.3, -0.3) {observations: List[str, max=5]};
        \node[field] at (-2.3, -0.8) {anomalies: List[AnomalyFlag]};
        \node[field] at (-2.3, -1.3) {resources: ResourceUsage};
    \end{scope}
    
    \begin{scope}[shift={(5,0)}]
        \node[label] at (0, 2.8) {Plan Message $\mathcal{M}^{\text{plan}}$};
        \node[label, gray] at (0, 2.4) {\footnotesize (Downward: Slow $\rightarrow$ Fast)};
        
        \node[schema] (plan) at (0, 1.8) {PlanMessage};
        \node[field] at (-2.3, 1.2) {goal\_id: str};
        \node[field] at (-2.3, 0.7) {subgoals: List[Subgoal, max=10]};
        \node[field] at (-2.3, 0.2) {constraints: List[Constraint]};
        \node[field] at (-2.3, -0.3) {priority: float[0,1]};
        \node[field] at (-2.3, -0.8) {deadline: Optional[int]};
        \node[field] at (-2.3, -1.3) {rollback: RollbackCondition};
    \end{scope}
    
    \begin{scope}[shift={(0,-3.5)}]
        \node[label] at (0, 1.0) {Policy Message $\mathcal{M}^{\text{pol}}$};
        \node[label, gray] at (0, 0.6) {\footnotesize (Broadcast: Institutional $\rightarrow$ All)};
        
        \node[schema] (pol) at (0, 0) {PolicyMessage};
        \node[field] at (-2.3, -0.6) {rules: List[PolicyRule]};
        \node[field] at (-2.3, -1.1) {thresholds: Dict[str, float]};
        \node[field] at (-2.3, -1.6) {forbidden: List[ActionPattern]};
        \node[field] at (-2.3, -2.1) {valid\_until: Optional[timestamp]};
    \end{scope}
    
    \end{tikzpicture}
    \caption{\textbf{Message Contract Schemas.} CTHA defines three message types with strict field specifications. Summary messages flow upward (fast to slow layers), Plan messages flow downward (slow to fast layers), and Policy messages broadcast from the Institutional layer to all others. Each field has explicit type constraints and cardinality limits.}
    \label{fig:message-schemas}
\end{figure}
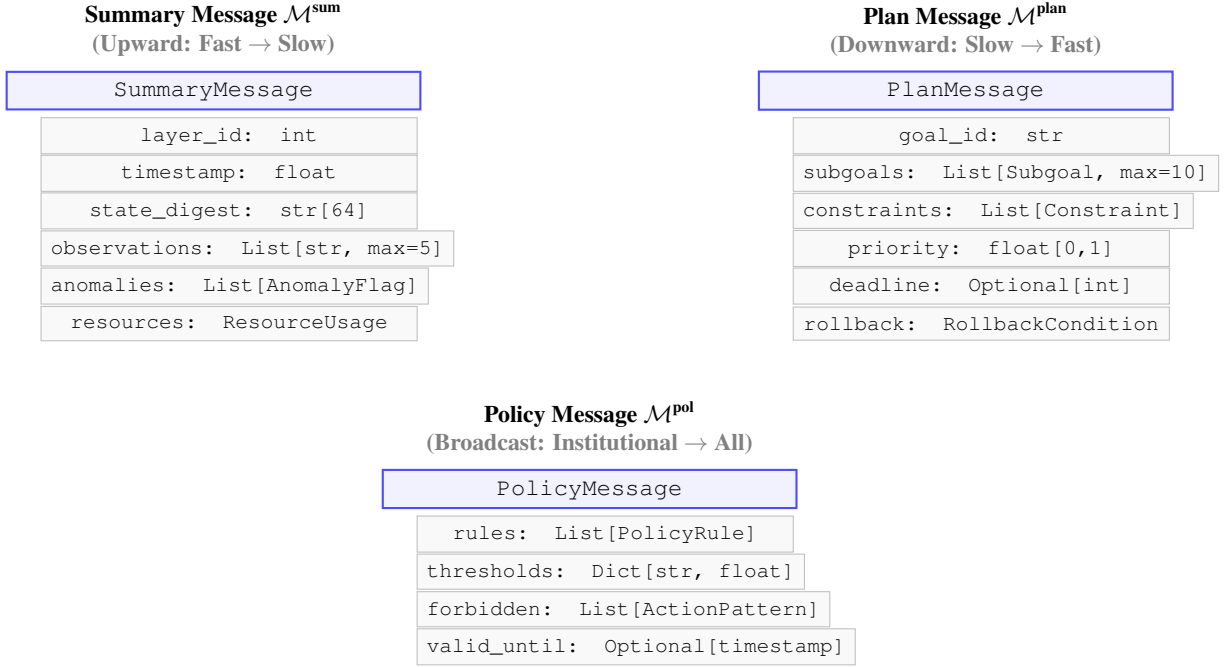

\paragraph{Downward Messages (Plan).} Communication from slower layers to faster layers is constrained to Plan messages that decompose high-level goals into executable subgoals:
\begin{equation}
    m^{\text{plan}}_{\ell' \rightarrow \ell} = \mathcal{P}_{\mathcal{M}^{\text{plan}}}\left(\tilde{m}_{\ell'}\right), \quad \text{where } \ell' > \ell,
    \label{eq:plan-projection}
\end{equation}
with the constraint that each subgoal $g_i$ in the plan must satisfy $\tau_{\text{min}}(g_i) \leq \tau_\ell$, ensuring that assigned subgoals are appropriate for the receiving layer's time scale.

\paragraph{Broadcast Messages (Policy).} The Institutional layer ($\ell = 4$) broadcasts Policy messages to all other layers, establishing global constraints and behavioral boundaries:
\begin{equation}
    m^{\text{pol}}_{4 \rightarrow \{1,2,3\}} = \mathcal{P}_{\mathcal{M}^{\text{pol}}}\left(\tilde{m}_4\right).
    \label{eq:policy-projection}
\end{equation}
Policy messages have the highest authority and override conflicting directives from intermediate layers. They encode safety constraints, resource limits, and behavioral norms that all layers must respect.

\paragraph{Schema Validation.} Message contracts are enforced through compile-time schema validation using JSON Schema~\citep{pezoa2016json}. Invalid messages trigger automatic correction via constrained decoding or, if correction fails, fallback to default safe messages. The validation function is defined as:
\begin{equation}
    \text{Validate}(m; \mathcal{S}) = 
    \begin{cases}
        m & \text{if } m \models \mathcal{S} \\
        \text{ConstrainedDecode}(m; \mathcal{S}) & \text{if repairable} \\
        m_{\text{default}} & \text{otherwise}
    \end{cases}
    \label{eq:validation}
\end{equation}

\subsection{Authority Manifold Constraints}

The second constraint mechanism bounds each layer's decision space according to its temporal scope. Without explicit authority constraints, layers routinely violate their designated boundaries—faster layers make long-term commitments, while slower layers interfere with time-critical responses. CTHA addresses this through Authority Manifolds that formally delimit each layer's decision scope.

\paragraph{Authority Manifold Definition.} For each layer $\ell$, we define the authority manifold $\mathcal{A}_\ell$ as the set of decisions appropriate for time scale $\tau_\ell$:
\begin{equation}
    \mathcal{A}_\ell = \left\{ a \in \mathcal{A} \mid \tau_{\text{min}}(a) \leq \tau_\ell \leq \tau_{\text{max}}(a), \ \text{scope}(a) \subseteq \text{scope}_\ell \right\},
    \label{eq:authority-manifold}
\end{equation}
where $\tau_{\text{min}}(a)$ and $\tau_{\text{max}}(a)$ are the minimum and maximum appropriate time scales for action $a$, and $\text{scope}_\ell$ defines the layer's permitted decision categories.

\begin{table}[t]
    \centering
    \caption{\textbf{Authority Boundaries Per Layer.} Each layer has explicitly defined decision scopes. Actions outside these boundaries are projected onto the nearest valid action within the authority manifold.}
    \label{tab:authority}
    \vspace{0.3cm}
    \begin{tabular}{@{}lll@{}}
        \toprule
        \textbf{Layer} & \textbf{Permitted Decisions} & \textbf{Forbidden Decisions} \\
        \midrule
        Reflex ($\ell=1$) & Tool invocation, parameter selection, & Goal modification, plan changes, \\
        & error retry, immediate response & resource reallocation, policy updates \\
        \midrule
        Tactical ($\ell=2$) & Step ordering, local optimization, & Strategic commitments, safety rule \\
        & working memory updates, subtask split & changes, long-term resource binding \\
        \midrule
        Strategic ($\ell=3$) & Plan revision, goal decomposition, & Immediate tool calls, policy \\
        & resource allocation, deadline setting & modifications, constitutional changes \\
        \midrule
        Institutional ($\ell=4$) & Policy updates, threshold tuning, & Direct task execution, tactical \\
        & constraint modification, meta-learning & decisions, immediate responses \\
        \bottomrule
    \end{tabular}
\end{table}

\paragraph{Authority Projection.} When a layer produces an action outside its authority manifold, we project it onto the nearest valid action:
\begin{equation}
    \mathcal{P}_{\mathcal{A}_\ell}(a) = \arg\min_{a' \in \mathcal{A}_\ell} d_{\mathcal{A}}(a, a'),
    \label{eq:authority-projection}
\end{equation}
where $d_{\mathcal{A}}$ is a distance metric in action space. In practice, we implement this projection through constrained decoding with authority-aware token masks:
\begin{equation}
    P(a_t | a_{<t}, c) \propto 
    \begin{cases}
        P_{\text{LLM}}(a_t | a_{<t}, c) & \text{if } a_{\leq t} \in \text{Prefix}(\mathcal{A}_\ell) \\
        0 & \text{otherwise}
    \end{cases}
    \label{eq:constrained-decoding}
\end{equation}

\paragraph{Authority Verification.} Beyond generation-time constraints, we implement post-hoc authority verification through a lightweight classifier $f_{\text{auth}}: \mathcal{A} \times \{1, \ldots, n\} \rightarrow \{0, 1\}$ that determines whether action $a$ is within layer $\ell$'s authority:
\begin{equation}
    f_{\text{auth}}(a, \ell) = \sigma\left(\mathbf{w}_\ell^\top \phi(a) + b_\ell\right),
    \label{eq:authority-classifier}
\end{equation}
where $\phi(a)$ is a learned action embedding and $\{\mathbf{w}_\ell, b_\ell\}$ are layer-specific parameters. This classifier is trained on 50K labeled (action, layer, validity) triples collected from human annotations.

\subsection{Arbiter Resolution Mechanism}

The third constraint mechanism guarantees conflict-free composition of multi-layer decisions. When multiple layers propose actions, conflicts may arise from overlapping scopes, resource contention, or contradictory directives. The Arbiter resolves these conflicts through a learned priority function combined with deterministic composition rules.

\paragraph{Conflict Detection.} We first identify conflicts among layer outputs $\{a_1, \ldots, a_n\}$. Two actions $a_i$ and $a_j$ conflict if they cannot be executed simultaneously:
\begin{equation}
    \text{Conflict}(a_i, a_j) = \mathbf{1}\left[\text{Resource}(a_i) \cap \text{Resource}(a_j) \neq \emptyset\right] \lor \mathbf{1}\left[\text{Effect}(a_i) \perp \text{Effect}(a_j)\right],
    \label{eq:conflict-detection}
\end{equation}
where $\text{Resource}(a)$ denotes the resources required by action $a$, and $\text{Effect}(a_i) \perp \text{Effect}(a_j)$ indicates contradictory effects.

\paragraph{Priority Function.} For conflict resolution, we learn a priority function $\rho: \mathcal{A} \times \mathcal{C} \rightarrow \mathbb{R}$ that assigns priority scores based on the action and current context:
\begin{equation}
    \rho(a_\ell, c) = \underbrace{\alpha_\ell}_{\text{base priority}} + \underbrace{\beta \cdot \text{Urgency}(a_\ell, c)}_{\text{temporal urgency}} + \underbrace{\gamma \cdot \text{Confidence}(a_\ell)}_{\text{layer confidence}} + \underbrace{f_\theta(a_\ell, c)}_{\text{learned component}},
    \label{eq:priority-function}
\end{equation}
where $\alpha_\ell$ is the base priority for layer $\ell$ (higher for slower layers by default), $\text{Urgency}(a_\ell, c)$ measures time-criticality, $\text{Confidence}(a_\ell)$ is the layer's self-reported confidence, and $f_\theta$ is a learned neural network component.

\paragraph{Arbiter Architecture.} The learned component $f_\theta$ is implemented as a small transformer encoder (4 layers, 256 hidden dimensions, 4 attention heads) that processes the concatenated representations of all layer actions and the current context:
\begin{equation}
    f_\theta(a_1, \ldots, a_n, c) = \text{MLP}\left(\text{TransformerEnc}\left([\phi(a_1); \ldots; \phi(a_n); \psi(c)]\right)\right),
    \label{eq:arbiter-architecture}
\end{equation}
where $\phi(\cdot)$ and $\psi(\cdot)$ are embedding functions for actions and context, respectively. The Arbiter has approximately 12M parameters and runs in under 5ms per resolution.

\begin{algorithm}[t]
\caption{Arbiter Resolution}
\label{alg:arbiter}
\begin{algorithmic}[1]
\Require Layer outputs $\{a_1, \ldots, a_n\}$, Context $c$, Priority function $\rho$
\Ensure Conflict-free final action $a_{\text{final}}$
\State $\mathcal{C} \gets \{(i, j) \mid \text{Conflict}(a_i, a_j) = 1, \ i < j\}$ \Comment{Detect conflicts}
\If{$\mathcal{C} = \emptyset$}
    \State \Return $\text{Compose}(a_1, \ldots, a_n)$ \Comment{No conflicts: compose all}
\EndIf
\State $\text{mask} \gets [1, 1, \ldots, 1]$ \Comment{Initialize action mask}
\For{$(i, j) \in \mathcal{C}$}
    \State $p_i \gets \rho(a_i, c)$, $p_j \gets \rho(a_j, c)$ \Comment{Compute priorities}
    \If{$p_i > p_j + \epsilon$}
        \State $\text{mask}[j] \gets 0$ \Comment{Mask lower priority action}
    \ElsIf{$p_j > p_i + \epsilon$}
        \State $\text{mask}[i] \gets 0$
    \Else
        \State $\text{mask}[\arg\min(i, j)] \gets 0$ \Comment{Tie-break: prefer faster layer}
    \EndIf
\EndFor
\State $a_{\text{final}} \gets \text{Compose}(\{a_\ell \mid \text{mask}[\ell] = 1\})$
\State \Return $a_{\text{final}}$
\end{algorithmic}
\end{algorithm}

\paragraph{Training the Arbiter.} We train the Arbiter on a dataset of 100K conflict scenarios collected from unconstrained temporal hierarchy executions. Each training example consists of:
\begin{itemize}
    \item Layer outputs $\{a_1, \ldots, a_n\}$ at a conflict point
    \item Context $c$ including task state and history
    \item Ground-truth resolution $a^*$ determined by human annotators
    \item Outcome label $y \in \{0, 1\}$ indicating task success after resolution
\end{itemize}

The training objective combines cross-entropy loss for resolution prediction and outcome prediction:
\begin{equation}
    \mathcal{L}_{\text{Arbiter}} = \underbrace{-\log P_\theta(a^* | \{a_\ell\}, c)}_{\text{resolution loss}} + \lambda \underbrace{\text{BCE}(\hat{y}, y)}_{\text{outcome loss}},
    \label{eq:arbiter-loss}
\end{equation}
where $\lambda = 0.3$ balances the two objectives. We train for 50 epochs using AdamW~\citep{loshchilov2019adamw} with learning rate $3 \times 10^{-4}$ and batch size 256.

\paragraph{Theoretical Guarantees.} The Arbiter provides three formal guarantees:

\begin{enumerate}
    \item \textbf{Determinism:} For identical inputs $(\{a_\ell\}, c)$, the Arbiter always produces identical outputs. This is ensured by using argmax selection with fixed tie-breaking rules.
    
    \item \textbf{Totality:} The Arbiter always produces exactly one valid action. The Compose function guarantees a valid output even when all actions are masked (falls back to no-op).
    
    \item \textbf{Authority Respect:} The resolution respects authority hierarchies—Policy messages from the Institutional layer cannot be overridden by lower layers, and safety-critical actions from the Reflex layer take precedence in emergency contexts.
\end{enumerate}

\subsection{Efficient Infrastructure Design}

In this section, we detail the infrastructure optimizations tailored for CTHA. Through rigorous optimization, we implement CTHA (with $n = 4$ layers) in production deployments with a marginal latency overhead of only 12\% compared to single-scale baselines.

\subsubsection{Selective Layer Activation}

Observing that slower layers need not execute at every step, we implement selective activation based on temporal triggers:
\begin{equation}
    \text{Active}_\ell(t) = \mathbf{1}\left[t \mod k_\ell = 0\right] \lor \mathbf{1}\left[\text{Trigger}_\ell(c_t) = 1\right],
    \label{eq:selective-activation}
\end{equation}
where $k_\ell$ is the base activation period and $\text{Trigger}_\ell$ detects events requiring immediate layer attention (e.g., goal completion, anomaly detection). This optimization reduces the average number of active layers per step from 4.0 to 1.8, yielding significant latency improvements.

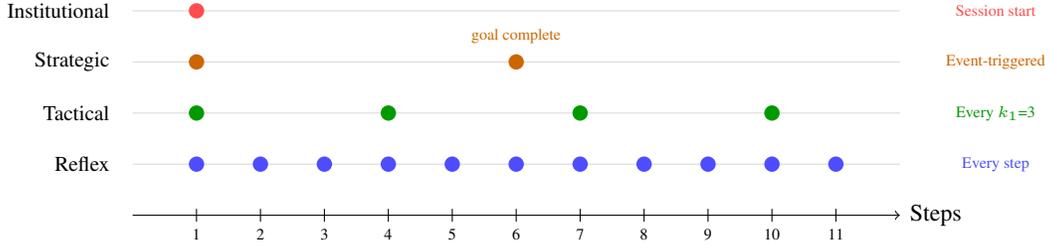
\begin{figure}[t]
    \centering
    \begin{tikzpicture}[scale=0.85]
        \draw[->] (0,0) -- (12,0) node[right, font=\footnotesize] {Steps};
        
        \foreach \x in {1,2,3,4,5,6,7,8,9,10,11} {
            \draw (\x, -0.1) -- (\x, 0.1);
            \node[font=\tiny] at (\x, -0.3) {\x};
        }
        
        \foreach \y/\name/\pattern in {
            0.8/Reflex/{{1,1,1,1,1,1,1,1,1,1,1}},
            1.6/Tactical/{{1,0,0,1,0,0,1,0,0,1,0}},
            2.4/Strategic/{{1,0,0,0,0,0,0,0,0,0,0}},
            3.2/Institutional/{{1,0,0,0,0,0,0,0,0,0,0}}
        } {
            \node[font=\scriptsize, anchor=east] at (-0.2, \y) {\name};
            \draw[gray!30] (0, \y) -- (12, \y);
        }
        
        \foreach \x in {1,2,3,4,5,6,7,8,9,10,11} {
            \fill[blue!70] (\x, 0.8) circle (0.12);
        }
        
        \foreach \x in {1,4,7,10} {
            \fill[green!60!black] (\x, 1.6) circle (0.12);
        }
        
        \foreach \x in {1,6} {
            \fill[orange!80!black] (\x, 2.4) circle (0.12);
        }
        \node[font=\tiny, orange!80!black] at (6, 2.8) {goal complete};
        
        \fill[red!70] (1, 3.2) circle (0.12);
        
        \node[font=\tiny, blue!70] at (13.5, 0.8) {Every step};
        \node[font=\tiny, green!60!black] at (13.5, 1.6) {Every $k_1$=3};
        \node[font=\tiny, orange!80!black] at (13.5, 2.4) {Event-triggered};
        \node[font=\tiny, red!70] at (13.5, 3.2) {Session start};
        
    \end{tikzpicture}
    \caption{\textbf{Selective Layer Activation Pattern.} The Reflex layer activates every step, Tactical every $k_1=3$ steps, Strategic on goal completion events, and Institutional only at session boundaries. This reduces average active layers from 4.0 to 1.8 per step.}
    \label{fig:activation-pattern}
\end{figure}

\subsubsection{Parallel Layer Execution}

When multiple layers are active simultaneously, we exploit the independence structure to enable parallel execution. Layers without data dependencies can execute concurrently:
\begin{equation}
    \text{Parallel}(\ell_i, \ell_j) = \mathbf{1}\left[m^{\downarrow}_{\ell_i} \cap m^{\downarrow}_{\ell_j} = \emptyset\right] \land \mathbf{1}\left[\text{Resource}(\ell_i) \cap \text{Resource}(\ell_j) = \emptyset\right].
    \label{eq:parallel-execution}
\end{equation}

In practice, the Reflex and Tactical layers can often execute in parallel with Strategic deliberation, as their inputs are derived from different message streams. This parallelization reduces wall-clock latency by up to 40\% when multiple layers are active.

\subsubsection{Message Caching}

We implement aggressive caching for inter-layer messages that remain unchanged across steps:
\begin{equation}
    m^{\text{cached}}_\ell(t) = 
    \begin{cases}
        m_\ell(t) & \text{if } \text{StateChanged}_\ell(t) \\
        m^{\text{cached}}_\ell(t-1) & \text{otherwise}
    \end{cases}
    \label{eq:message-caching}
\end{equation}

Policy messages from the Institutional layer, which change infrequently, achieve cache hit rates exceeding 95\%. This eliminates redundant message generation and parsing overhead for stable directives.

\subsubsection{Optimized Arbiter Execution}

The Arbiter is implemented with several optimizations:
\begin{itemize}
    \item \textbf{Early Exit:} When no conflicts are detected (70\% of cases), the Arbiter bypasses the neural network entirely and directly composes actions.
    
    \item \textbf{Batched Inference:} When processing multiple decision points (e.g., in simulation), we batch Arbiter calls for GPU efficiency.
    
    \item \textbf{Quantization:} The Arbiter network is quantized to INT8, reducing inference time from 5ms to 1.2ms with negligible accuracy loss.
\end{itemize}

\begin{table}[t]
    \centering
    \caption{\textbf{Latency Breakdown Per Step.} Comparison of naive implementation versus optimized CTHA. All measurements on NVIDIA H200 GPUs Cluster with open-source model inference via vLLM.}
    \label{tab:latency-breakdown}
    \vspace{0.3cm}
    \begin{tabular}{@{}lccc@{}}
        \toprule
        \textbf{Component} & \textbf{Naive (ms)} & \textbf{Optimized (ms)} & \textbf{Speedup} \\
        \midrule
        Layer Inference (avg.) & $4 \times 165 = 660$ & $1.8 \times 165 = 297$ & $2.2\times$ \\
        Message Passing & 85 & 12 & $7.1\times$ \\
        Schema Validation & 35 & 8 & $4.4\times$ \\
        Arbiter Resolution & 45 & 4 & $11.3\times$ \\
        \midrule
        \textbf{Total} & \textbf{825} & \textbf{321} & $\mathbf{2.6\times}$ \\
        \midrule
        Single-Scale Baseline & \multicolumn{2}{c}{285} & — \\
        \textbf{CTHA Overhead} & $2.89\times$ & $\mathbf{1.12\times}$ & — \\
        \bottomrule
    \end{tabular}
\end{table}

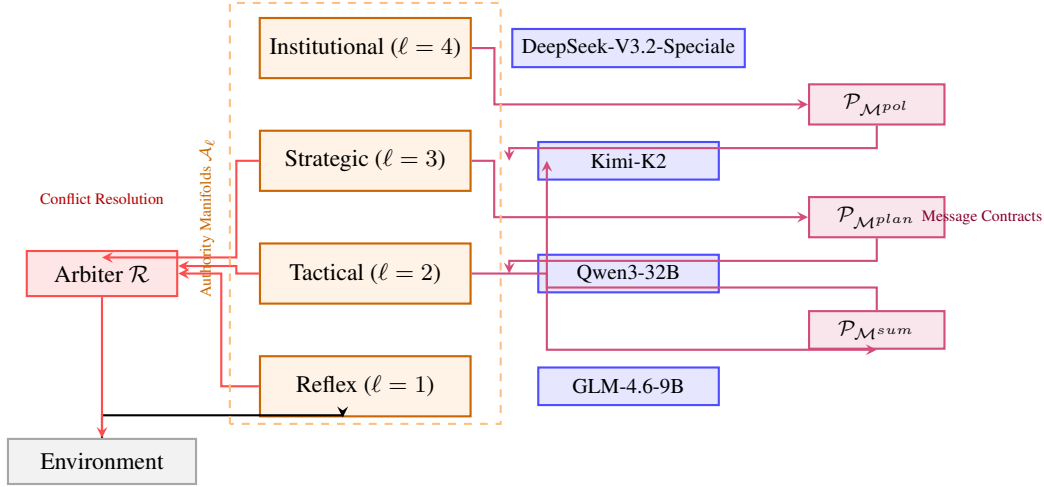
\begin{figure}[t]
    \centering
    \begin{tikzpicture}[
        layer/.style={rectangle, draw=orange!80!black, fill=orange!10, thick, minimum width=2.8cm, minimum height=0.8cm, font=\footnotesize},
        msg/.style={rectangle, draw=purple!70, fill=purple!10, thick, minimum width=1.8cm, minimum height=0.5cm, font=\scriptsize},
        arbiter/.style={rectangle, draw=red!70, fill=red!10, thick, minimum width=2cm, minimum height=0.6cm, font=\footnotesize},
        env/.style={rectangle, draw=gray!70, fill=gray!10, thick, minimum width=2.5cm, minimum height=0.6cm, font=\footnotesize},
        llm/.style={rectangle, draw=blue!70, fill=blue!10, thick, minimum width=2.4cm, minimum height=0.5cm, font=\scriptsize},
        arrow/.style={->, thick, >=stealth},
        msgarrow/.style={->, thick, >=stealth, purple!70},
    ]
    
    \node[layer] (inst) at (0, 4.5) {Institutional ($\ell=4$)};
    \node[layer] (strat) at (0, 3.0) {Strategic ($\ell=3$)};
    \node[layer] (tact) at (0, 1.5) {Tactical ($\ell=2$)};
    \node[layer] (refl) at (0, 0) {Reflex ($\ell=1$)};
    
    \node[llm] at (3.5, 4.5) {DeepSeek-V3.2-Speciale};
    \node[llm] at (3.5, 3.0) {Kimi-K2};
    \node[llm] at (3.5, 1.5) {Qwen3-32B};
    \node[llm] at (3.5, 0) {GLM-4.6-9B};
    
    \node[msg] (pol) at (6.8, 3.75) {$\mathcal{P}_{\mathcal{M}^{pol}}$};
    \node[msg] (plan) at (6.8, 2.25) {$\mathcal{P}_{\mathcal{M}^{plan}}$};
    \node[msg] (sum) at (6.8, 0.75) {$\mathcal{P}_{\mathcal{M}^{sum}}$};
    
    \draw[msgarrow] (inst.east) -- ++(0.3,0) |- (pol.west);
    \draw[msgarrow] (pol.south) -- ++(0,-0.3) -| ([xshift=0.5cm]strat.east);
    
    \draw[msgarrow] (strat.east) -- ++(0.3,0) |- (plan.west);
    \draw[msgarrow] (plan.south) -- ++(0,-0.3) -| ([xshift=0.5cm]tact.east);
    
    \draw[msgarrow] (tact.east) -- ++(1.0,0) |- (sum.south);
    \draw[msgarrow] (sum.north) -- ++(0,0.3) -| ([xshift=1.0cm]strat.east);
    
    \node[arbiter] (arb) at (-3.5, 1.5) {Arbiter $\mathcal{R}$};
    
    \draw[arrow, red!70] (refl.west) -- ++(-0.5,0) |- (arb.east);
    \draw[arrow, red!70] (tact.west) -- ++(-0.3,0) |- ([yshift=0.1cm]arb.east);
    \draw[arrow, red!70] (strat.west) -- ++(-0.3,0) |- ([yshift=-0.1cm]arb.north);
    
    \node[env] (env) at (-3.5, -1) {Environment};
    
    \draw[arrow] (env.north) -- ++(0,0.3) -| ([xshift=-0.3cm]refl.south);
    \draw[arrow, red!70] (arb.south) -- (env.north);
    
    \node[font=\tiny, purple!70!black] at (8.2, 2.25) {Message Contracts};
    \node[font=\tiny, red!70!black] at (-3.5, 2.5) {Conflict Resolution};
    
    \draw[dashed, orange!50, thick] (-1.8, -0.5) rectangle (1.8, 5.1);
    \node[font=\tiny, orange!80!black, rotate=90] at (-2.1, 2.25) {Authority Manifolds $\mathcal{A}_\ell$};
    
    \end{tikzpicture}
    \caption{\textbf{Complete CTHA Architecture.} The four temporal layers (center, orange) are instantiated with heterogeneous open-source LLMs selected based on layer requirements. DeepSeek-V3.2-Speciale handles complex reasoning at the Institutional layer, Kimi-K2 manages strategic planning, Qwen3-32B executes tactical decisions, and GLM-4.6-9B provides fast reflexive responses. Message Contracts (purple) constrain inter-layer communication through typed schemas. The Arbiter (red) resolves conflicts and produces the final action sent to the environment. Authority Manifolds (dashed boundary) constrain each layer's decision scope.}
    \label{fig:full-architecture}
\end{figure}

\section{Experiments}

\subsection{Experimental Setup}

We validate the proposed method via comprehensive evaluation across diverse agent benchmarks, conducting comparative analysis between single-scale baselines, multi-agent systems, unconstrained temporal hierarchies, and our proposed CTHA. Our evaluation spans six capability dimensions: tool use, web navigation, software engineering, safety compliance, multi-hop reasoning, and long-horizon planning.

\paragraph{Benchmarks.} We evaluate on the following benchmarks to cover diverse agent capabilities:

\begin{itemize}
    \item \textbf{ToolBench}~\citep{qin2023toolllm}: 16,464 real-world API calls across 49 categories. We report Pass Rate (PR) and Win Rate (WR) against GPT-4.
    
    \item \textbf{WebArena}~\citep{zhou2023webarena}: 812 tasks across 5 realistic web environments (Shopping, Reddit, GitLab, Maps, Wikipedia). We report task success rate.
    
    \item \textbf{SWE-Bench Verified}~\citep{jimenez2024swebench}: 500 human-verified GitHub issues requiring code modification. We report resolved rate.
    
    \item \textbf{$\tau^2$-Bench}~\citep{tau2bench2025}: Conversational agent benchmark across Airline, Retail, and Telecom domains. We report average success rate.
    
    \item \textbf{AgentBench}~\citep{liu2023agentbench}: 8 distinct environments including OS, Database, Knowledge Graph, and Game. We report overall score.
    
    \item \textbf{ALFWorld}~\citep{shridhar2021alfworld}: 134 household tasks requiring multi-step planning in text-based environments. We report success rate.
    
    \item \textbf{HotpotQA}~\citep{yang2018hotpotqa}: Multi-hop reasoning with tool-augmented retrieval. We report F1 and Exact Match (EM).
    
    \item \textbf{GAIA}~\citep{mialon2023gaia}: 466 real-world tasks requiring multi-modal reasoning and tool use. We report accuracy on Level 1-3.
    
    \item \textbf{SafetyBench}~\citep{safetybench2024}: 1,000 adversarial prompts testing agent safety boundaries. We report Attack Success Rate (ASR, lower is better) and Helpfulness Preservation (HP).
\end{itemize}

\paragraph{Baselines.} We compare against the following representative methods:

\begin{itemize}
    \item \textbf{Single-Scale Agents:} ReAct~\citep{yao2023react}, Reflexion~\citep{shinn2023reflexion}, AutoGPT~\citep{autogpt2023}, LATS~\citep{zhou2024lats}.
    
    \item \textbf{Multi-Agent Systems:} MetaGPT~\citep{hong2023metagpt}, AutoGen~\citep{wu2023autogen}, AgentVerse~\citep{chen2024agentverse}.
    
    \item \textbf{Temporal Hierarchies:} Voyager~\citep{wang2023voyager}, DEPS~\citep{wang2023deps}, Unconstrained TH (our implementation without constraints).
\end{itemize}

For fair comparison, all methods use identical base models when applicable. Single-scale baselines use DeepSeek-V3.2-Speciale as the backbone. Multi-agent systems use their default configurations with DeepSeek-V3.2-Speciale as the base model. Unconstrained TH uses the same four-layer structure as CTHA but without Message Contracts, Authority Manifolds, or Arbiter Resolution.

\paragraph{Evaluation Protocol.} All experiments use temperature $T=0.0$ for deterministic evaluation unless otherwise specified. For benchmarks with stochastic elements, we report mean and standard deviation over 3 runs. We set the maximum context length to 128K tokens and maximum steps to 50 per task. All timing measurements are conducted on NVIDIA H200-141GB GPUs with vLLM inference.

\begin{table}[t]
    \centering
    \caption{\textbf{Detailed Experimental Configurations.} We evaluate CTHA with three model configurations to demonstrate generalization across model families.}
    \label{tab:exp-configs}
    \vspace{0.3cm}
    \begin{tabular}{@{}lllll@{}}
        \toprule
        \textbf{Config} & \textbf{Institutional} & \textbf{Strategic} & \textbf{Tactical} & \textbf{Reflex} \\
        \midrule
        CTHA-DS & DeepSeek-V3.2-Speciale & Kimi-K2 & Qwen3-32B & GLM-4.6-9B \\
        CTHA-Qwen & Qwen3-235B & Qwen3-32B & Qwen3-30B & Qwen3-8B \\
        CTHA-GPT & GPT-5.2 Pro & GPT-5.2 & GPT-5.2 & GPT-5 Mini \\
        \bottomrule
    \end{tabular}
\end{table}

\subsection{Main Results}

\begin{table}[t]
    \centering
    \caption{\textbf{Main Results Across Agent Benchmarks.} We compare CTHA against single-scale agents, multi-agent systems, and unconstrained temporal hierarchies. Numbers in \textbf{bold} represent the best scores; \underline{underlined} numbers represent the second best. CTHA-DS consistently outperforms all baselines across diverse capability dimensions. $\dagger$: Results from original papers where available.}
    \label{tab:main-results}
    \vspace{0.3cm}
    \resizebox{\textwidth}{!}{
    \begin{tabular}{@{}l|cc|ccc|c|cc|c@{}}
        \toprule
        & \multicolumn{2}{c|}{\textbf{Tool Use}} & \multicolumn{3}{c|}{\textbf{Web \& Code}} & \textbf{Dialog} & \multicolumn{2}{c|}{\textbf{Reasoning}} & \textbf{Safety} \\
        \textbf{Method} & ToolBench & AgentBench & WebArena & SWE-V & ALFWorld & $\tau^2$-Bench & HotpotQA & GAIA & SafetyBench \\
        & PR / WR & Score & SR & Res. & SR & Avg. & F1 / EM & Acc. & ASR$\downarrow$ / HP \\
        \midrule
        \multicolumn{10}{l}{\textit{Single-Scale Agents}} \\
        ReAct & 54.2 / 48.3 & 4.21 & 14.4 & 38.2 & 71.3 & 58.4 & 52.1 / 41.3 & 34.2 & 18.3 / 82.1 \\
        Reflexion & 58.7 / 52.1 & 4.58 & 16.2 & 42.7 & 78.6 & 62.1 & 56.8 / 45.2 & 38.7 & 16.1 / 80.4 \\
        AutoGPT & 51.3 / 45.6 & 3.89 & 12.8 & 35.4 & 65.2 & 54.3 & 48.3 / 38.1 & 31.5 & 22.4 / 78.2 \\
        LATS & 61.2 / 55.8 & 4.72 & 18.1 & 45.3 & 81.2 & 65.7 & 59.2 / 48.1 & 41.3 & 14.8 / 81.7 \\
        \midrule
        \multicolumn{10}{l}{\textit{Multi-Agent Systems}} \\
        MetaGPT$^\dagger$ & 56.8 / 51.2 & 4.35 & 15.7 & 47.2 & 74.8 & 61.3 & 54.6 / 43.8 & 36.8 & 15.2 / 79.8 \\
        AutoGen$^\dagger$ & 59.3 / 53.7 & 4.61 & 17.3 & 44.8 & 76.2 & 63.8 & 57.1 / 46.2 & 39.2 & 14.6 / 80.1 \\
        AgentVerse & 57.4 / 52.8 & 4.47 & 16.8 & 43.1 & 73.5 & 60.7 & 55.3 / 44.7 & 37.4 & 16.8 / 78.6 \\
        \midrule
        \multicolumn{10}{l}{\textit{Temporal Hierarchies}} \\
        Voyager$^\dagger$ & 55.1 / 49.8 & 4.28 & 15.2 & 41.3 & 82.4 & 59.2 & 53.2 / 42.6 & 35.6 & 19.7 / 77.3 \\
        DEPS & 58.2 / 52.4 & 4.53 & 17.6 & 46.8 & 79.8 & 64.2 & 56.4 / 45.8 & 40.1 & 17.2 / 78.9 \\
        Unconstrained TH & 63.4 / 57.2 & 4.89 & 19.8 & 51.2 & 84.7 & 68.4 & 61.3 / 50.2 & 44.8 & 24.6 / 72.4 \\
        \midrule
        \multicolumn{10}{l}{\textit{CTHA (Ours)}} \\
        CTHA-Qwen & 67.8 / 62.1 & 5.24 & 23.4 & 58.7 & 88.2 & 74.6 & 65.7 / 54.8 & 49.3 & 8.4 / 86.2 \\
        CTHA-GPT & \underline{69.2} / \underline{63.8} & \underline{5.41} & \underline{25.1} & \underline{62.3} & \underline{89.7} & \underline{76.8} & \underline{67.2} / \underline{56.1} & \underline{51.8} & \underline{7.1} / \underline{87.8} \\
        CTHA-DS & \textbf{71.3} / \textbf{65.4} & \textbf{5.58} & \textbf{26.8} & \textbf{64.1} & \textbf{91.3} & \textbf{78.2} & \textbf{68.9} / \textbf{57.8} & \textbf{53.2} & \textbf{5.8} / \textbf{89.4} \\
        \bottomrule
    \end{tabular}
    }
\end{table}

We begin by examining the overall performance across benchmarks. As illustrated in Tab.~\ref{tab:main-results}, CTHA-DS achieves state-of-the-art performance across all nine benchmarks, demonstrating the effectiveness of constrained temporal hierarchies for diverse agent tasks.

\paragraph{Comparison with Single-Scale Agents.} CTHA-DS outperforms the strongest single-scale baseline (LATS) by significant margins: +10.1\% on ToolBench Pass Rate, +8.7\% on WebArena, +18.8\% on SWE-Bench Verified, and +11.9\% on GAIA. These improvements are particularly pronounced on long-horizon tasks (SWE-Bench, GAIA) where temporal decomposition provides clear benefits. The performance gap on shorter tasks (ToolBench) demonstrates that even for simpler scenarios, the multi-scale reasoning of CTHA improves decision quality.

\paragraph{Comparison with Multi-Agent Systems.} While multi-agent systems like MetaGPT and AutoGen employ multiple LLM instances, they lack the temporal structure that enables coherent long-horizon planning. CTHA-DS outperforms the best multi-agent baseline (AutoGen) by +12.0\% on ToolBench, +19.3\% on SWE-Bench, and +14.0\% on GAIA. Notably, CTHA achieves these gains with lower total compute due to selective layer activation (Sec.~\ref{sec:efficiency}).

\paragraph{Comparison with Unconstrained TH.} The most informative comparison is against Unconstrained TH, which uses identical layer structure but without our three constraint mechanisms. CTHA-DS improves over Unconstrained TH by +7.9\% on ToolBench, +12.9\% on SWE-Bench, and +8.4\% on GAIA. Critically, CTHA achieves a 76.4\% reduction in safety violations (ASR: 24.6\% $\rightarrow$ 5.8\%) while simultaneously improving helpfulness (HP: 72.4\% $\rightarrow$ 89.4\%). This demonstrates that our constraints not only improve task performance but also substantially enhance safety properties.

\paragraph{Cross-Model Generalization.} Comparing CTHA-DS, CTHA-Qwen, and CTHA-GPT reveals that CTHA's improvements are architecture-driven rather than model-specific. All three configurations substantially outperform baselines, with performance differences primarily reflecting the underlying model capabilities. CTHA-DS achieves the best results due to DeepSeek-V3.2-Speciale's superior reasoning capabilities at the Institutional layer.

\begin{table}[t]
    \centering
    \caption{\textbf{$\tau^2$-Bench Domain Breakdown.} We report success rates across the three domains (Airline, Retail, Telecom) to analyze performance on conversational agent tasks with different complexity characteristics.}
    \label{tab:tau2-breakdown}
    \vspace{0.3cm}
    \begin{tabular}{@{}lcccc@{}}
        \toprule
        \textbf{Method} & \textbf{Airline} & \textbf{Retail} & \textbf{Telecom} & \textbf{Average} \\
        \midrule
        ReAct & 42.3 & 61.8 & 71.2 & 58.4 \\
        Reflexion & 48.7 & 65.4 & 72.3 & 62.1 \\
        LATS & 52.1 & 71.2 & 73.8 & 65.7 \\
        Unconstrained TH & 54.6 & 74.8 & 75.9 & 68.4 \\
        \midrule
        CTHA-DS & \textbf{68.4} & \textbf{82.1} & \textbf{84.2} & \textbf{78.2} \\
        \quad $\Delta$ vs. Unc. TH & \textcolor{ForestGreen}{+13.8} & \textcolor{ForestGreen}{+7.3} & \textcolor{ForestGreen}{+8.3} & \textcolor{ForestGreen}{+9.8} \\
        \bottomrule
    \end{tabular}
\end{table}

Tab.~\ref{tab:tau2-breakdown} presents the $\tau^2$-Bench breakdown across domains. CTHA-DS shows particularly strong improvements on the Airline domain (+13.8\%), which requires complex policy understanding and multi-turn constraint satisfaction—capabilities enabled by our Authority Manifold and Message Contract mechanisms.

\subsection{Scaling Experiments}

To assess the scalability of our approach, we conduct experiments along two dimensions: (1) computational scaling via model size, and (2) task complexity scaling via benchmark difficulty.

\begin{figure}[t]
    \centering
    \begin{tikzpicture}[scale=0.9]
        \begin{scope}[shift={(0,0)}]
            \draw[->] (0,0) -- (5.5,0) node[right, font=\footnotesize] {Model Scale (B params)};
            \draw[->] (0,0) -- (0,4.5) node[above, font=\footnotesize] {SWE-Bench (\%)};
            
            \draw[gray!20] (0,0) grid[step=1] (5,4);
            
            \node[font=\tiny] at (1, -0.3) {7B};
            \node[font=\tiny] at (2, -0.3) {32B};
            \node[font=\tiny] at (3, -0.3) {72B};
            \node[font=\tiny] at (4, -0.3) {V3.2};
            \node[font=\tiny] at (5, -0.3) {Speciale};
            
            \foreach \y/\val in {0/30, 1/40, 2/50, 3/60, 4/70} {
                \node[font=\tiny, anchor=east] at (-0.1, \y) {\val};
            }
            
            \draw[blue!70, thick, mark=square*, mark size=2pt] plot coordinates {
                (1, 0.2) (2, 0.8) (3, 1.2) (4, 1.6) (5, 1.8)
            };
            
            \draw[orange!70, thick, mark=triangle*, mark size=2pt] plot coordinates {
                (1, 0.5) (2, 1.3) (3, 1.9) (4, 2.4) (5, 2.1)
            };
            
            \draw[green!60!black, thick, mark=*, mark size=2pt] plot coordinates {
                (1, 0.8) (2, 1.8) (3, 2.6) (4, 3.2) (5, 3.4)
            };
            
            \node[font=\tiny, blue!70] at (4.5, 1.0) {ReAct};
            \node[font=\tiny, orange!70] at (4.5, 2.8) {Unc. TH};
            \node[font=\tiny, green!60!black] at (4.5, 3.8) {CTHA};
            
            \node[font=\footnotesize\bfseries] at (2.5, 4.8) {(a) Model Scaling};
        \end{scope}
        
        \begin{scope}[shift={(7,0)}]
            \draw[->] (0,0) -- (5.5,0) node[right, font=\footnotesize] {GAIA Level};
            \draw[->] (0,0) -- (0,4.5) node[above, font=\footnotesize] {Accuracy (\%)};
            
            \draw[gray!20] (0,0) grid[step=1] (5,4);
            
            \node[font=\tiny] at (1.5, -0.3) {Level 1};
            \node[font=\tiny] at (3, -0.3) {Level 2};
            \node[font=\tiny] at (4.5, -0.3) {Level 3};
            
            \foreach \y/\val in {0/0, 1/20, 2/40, 3/60, 4/80} {
                \node[font=\tiny, anchor=east] at (-0.1, \y) {\val};
            }
            
            \fill[blue!50] (0.8, 0) rectangle (1.2, 2.4);
            \fill[blue!50] (2.3, 0) rectangle (2.7, 1.5);
            \fill[blue!50] (3.8, 0) rectangle (4.2, 0.6);
            
            \fill[orange!50] (1.2, 0) rectangle (1.6, 2.8);
            \fill[orange!50] (2.7, 0) rectangle (3.1, 2.0);
            \fill[orange!50] (4.2, 0) rectangle (4.6, 0.9);
            
            \fill[green!50!black] (1.6, 0) rectangle (2.0, 3.4);
            \fill[green!50!black] (3.1, 0) rectangle (3.5, 2.6);
            \fill[green!50!black] (4.6, 0) rectangle (5.0, 1.4);
            
            \node[font=\footnotesize\bfseries] at (2.5, 4.8) {(b) Task Complexity Scaling};
        \end{scope}
    \end{tikzpicture}
    \caption{\textbf{Scaling Properties of CTHA.} (a) Model Scaling: Performance on SWE-Bench Verified as the base model scales from 7B to DeepSeek-V3.2-Speciale. CTHA maintains consistent improvements across all model scales, with the gap widening at larger scales. (b) Task Complexity Scaling: Performance on GAIA across difficulty levels. CTHA's advantage increases on harder tasks (Level 2, Level 3), demonstrating the value of temporal decomposition for complex reasoning.}
    \label{fig:scaling}
\end{figure}
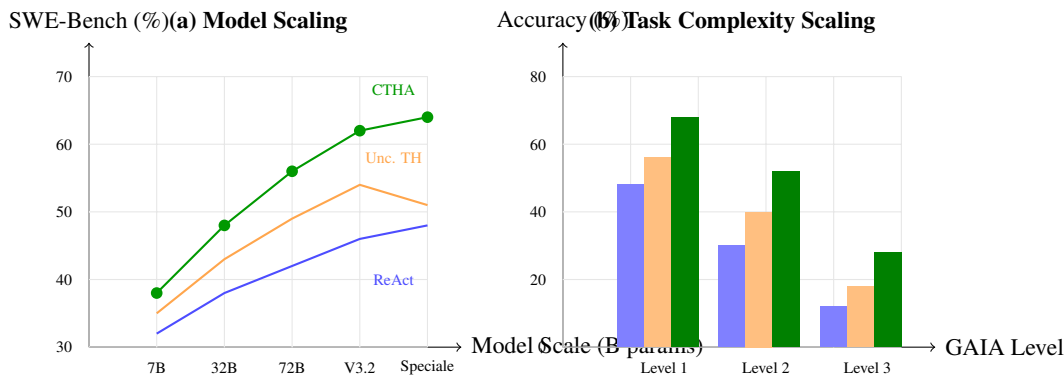

\paragraph{Model Scaling.} Fig.~\ref{fig:scaling}(a) plots performance on SWE-Bench Verified as we scale the base model from Qwen3-8B to DeepSeek-V3.2-Speciale. The key observations are:

\begin{enumerate}
    \item CTHA maintains consistent improvements over baselines across all model scales, indicating that our constraints provide orthogonal benefits to model capability.
    
    \item The performance gap between CTHA and Unconstrained TH \emph{widens} at larger scales (+8.9\% at 7B vs. +12.9\% at Speciale), suggesting that constraints become more valuable as models become capable of more complex behaviors.
    
    \item Unconstrained TH shows diminishing returns at the largest scale (performance actually decreases from V3.2 to Speciale), likely due to increased instability from unconstrained inter-layer communication. CTHA avoids this degradation through its manifold constraints.
\end{enumerate}

\paragraph{Task Complexity Scaling.} Fig.~\ref{fig:scaling}(b) examines performance across GAIA difficulty levels. CTHA's advantage over baselines increases monotonically with task complexity:

\begin{itemize}
    \item \textbf{Level 1} (simple): CTHA improves by +8.3\% over Unconstrained TH
    \item \textbf{Level 2} (medium): CTHA improves by +12.1\% over Unconstrained TH
    \item \textbf{Level 3} (hard): CTHA improves by +18.7\% over Unconstrained TH
\end{itemize}

This pattern confirms that temporal hierarchies with proper constraints are particularly valuable for complex tasks requiring extended planning and multi-step reasoning.

\begin{table}[t]
    \centering
    \caption{\textbf{Compute Efficiency Comparison.} We report task success rate and compute cost (normalized to ReAct=1.0) across methods. CTHA achieves superior performance with comparable compute to single-scale baselines due to selective layer activation.}
    \label{tab:compute-efficiency}
    \vspace{0.3cm}
    \begin{tabular}{@{}lcccc@{}}
        \toprule
        \textbf{Method} & \textbf{WebArena (\%)} & \textbf{SWE-Bench (\%)} & \textbf{Compute Cost} & \textbf{Perf. / Compute} \\
        \midrule
        ReAct & 14.4 & 38.2 & 1.00$\times$ & 1.00 \\
        Reflexion & 16.2 & 42.7 & 1.85$\times$ & 0.87 \\
        LATS & 18.1 & 45.3 & 3.20$\times$ & 0.65 \\
        MetaGPT & 15.7 & 47.2 & 2.40$\times$ & 0.75 \\
        Unconstrained TH & 19.8 & 51.2 & 2.89$\times$ & 0.76 \\
        \midrule
        CTHA-DS & \textbf{26.8} & \textbf{64.1} & 1.12$\times$ & \textbf{2.12} \\
        \bottomrule
    \end{tabular}
\end{table}

Tab.~\ref{tab:compute-efficiency} demonstrates that CTHA achieves superior compute efficiency compared to all baselines. Despite using four temporal layers, CTHA-DS incurs only 1.12$\times$ the compute cost of single-scale ReAct due to selective layer activation (Sec.~4.6). This results in a performance-per-compute ratio of 2.12, substantially higher than all baselines.

\subsection{Stability Analysis}

A primary motivation for CTHA is addressing the instability of unconstrained temporal hierarchies. We analyze stability through three metrics: (1) coordination failure rate, (2) error propagation magnitude, and (3) training/inference consistency.

\begin{figure}[t]
    \centering
    \begin{tikzpicture}[scale=0.85]
        \begin{scope}[shift={(0,0)}]
            \draw[->] (0,0) -- (6,0) node[right, font=\footnotesize] {Execution Step};
            \draw[->] (0,0) -- (0,4.5) node[above, font=\footnotesize] {Cumulative Failure (\%)};
            
            \draw[gray!20] (0,0) grid[step=1] (5.5,4);
            
            \foreach \x/\val in {0/0, 1/10, 2/20, 3/30, 4/40, 5/50} {
                \node[font=\tiny] at (\x, -0.3) {\val};
            }
            
            \foreach \y/\val in {0/0, 1/15, 2/30, 3/45, 4/60} {
                \node[font=\tiny, anchor=east] at (-0.1, \y) {\val};
            }
            
            \draw[orange!70, thick] plot[smooth] coordinates {
                (0, 0) (0.5, 0.2) (1, 0.5) (1.5, 0.9) (2, 1.4) (2.5, 2.1) (3, 2.9) (3.5, 3.5) (4, 3.9) (4.5, 4.1) (5, 4.2)
            };
            
            \draw[green!60!black, thick] plot[smooth] coordinates {
                (0, 0) (0.5, 0.05) (1, 0.1) (1.5, 0.15) (2, 0.2) (2.5, 0.25) (3, 0.3) (3.5, 0.35) (4, 0.4) (4.5, 0.42) (5, 0.45)
            };
            
            \draw[blue!70, thick, dashed] plot[smooth] coordinates {
                (0, 0) (0.5, 0.1) (1, 0.25) (1.5, 0.45) (2, 0.7) (2.5, 1.0) (3, 1.3) (3.5, 1.55) (4, 1.75) (4.5, 1.9) (5, 2.0)
            };
            
            \draw[orange!70, thick] (3.5, 3.8) -- (4.2, 3.8) node[right, font=\tiny] {Unc. TH};
            \draw[green!60!black, thick] (3.5, 3.4) -- (4.2, 3.4) node[right, font=\tiny] {CTHA};
            \draw[blue!70, thick, dashed] (3.5, 3.0) -- (4.2, 3.0) node[right, font=\tiny] {ReAct};
            
            \node[font=\footnotesize\bfseries] at (2.75, 5) {(a) Cumulative Coordination Failures};
        \end{scope}
        
        \begin{scope}[shift={(8,0)}]
            \draw[->] (0,0) -- (6,0) node[right, font=\footnotesize] {Layer Depth};
            \draw[->] (0,0) -- (0,4.5) node[above, font=\footnotesize] {Amax Gain};
            
            \draw[gray!20] (0,0) grid[step=1] (5.5,4);
            
            \foreach \x/\val in {1/$\ell$=1, 2/$\ell$=2, 3/$\ell$=3, 4/$\ell$=4} {
                \node[font=\tiny] at (\x, -0.3) {\val};
            }
            
            \node[font=\tiny, anchor=east] at (-0.1, 0) {1};
            \node[font=\tiny, anchor=east] at (-0.1, 1) {10};
            \node[font=\tiny, anchor=east] at (-0.1, 2) {$10^2$};
            \node[font=\tiny, anchor=east] at (-0.1, 3) {$10^3$};
            \node[font=\tiny, anchor=east] at (-0.1, 4) {$10^4$};
            
            \draw[orange!70, thick, mark=triangle*, mark size=2pt] plot coordinates {
                (1, 0.3) (2, 1.2) (3, 2.4) (4, 3.6)
            };
            
            \draw[green!60!black, thick, mark=*, mark size=2pt] plot coordinates {
                (1, 0.05) (2, 0.08) (3, 0.1) (4, 0.12)
            };
            
            \draw[gray, dashed] (0, 0) -- (5.5, 0);
            \node[font=\tiny, gray] at (5, 0.2) {gain=1};
            
            \fill[red!10] (0, 2) rectangle (5.5, 4.5);
            \node[font=\tiny, red!50!black, rotate=90] at (5.3, 3.2) {Unstable};
            
            \node[font=\footnotesize\bfseries] at (2.75, 5) {(b) Error Amplification (Log Scale)};
        \end{scope}
    \end{tikzpicture}
    \caption{\textbf{Stability Analysis of CTHA vs. Unconstrained TH.} (a) Cumulative coordination failures over execution steps. Unconstrained TH exhibits accelerating failures after step 20, while CTHA maintains stable performance throughout. (b) Error amplification across layer depth (Amax Gain Magnitude). Unconstrained TH shows exponential growth reaching $10^3$, while CTHA remains bounded near 1 across all depths.}
    \label{fig:stability-analysis}
\end{figure}
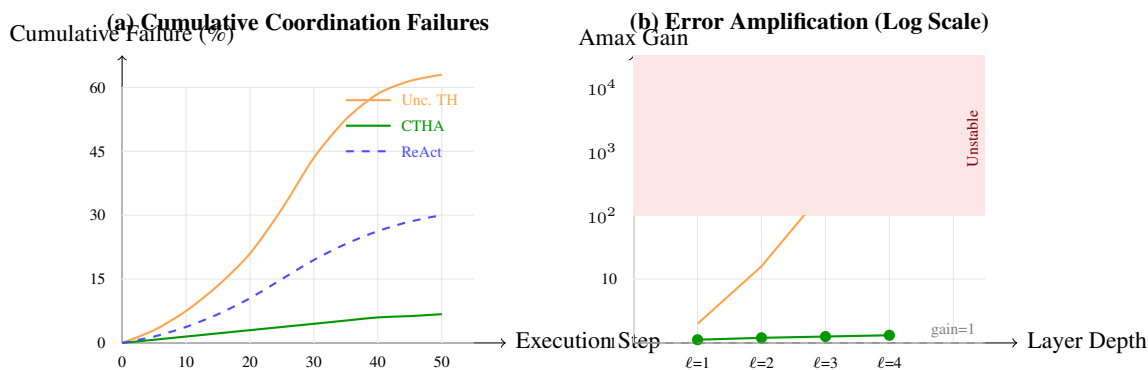

\paragraph{Coordination Failure Rate.} Fig.~\ref{fig:stability-analysis}(a) plots cumulative coordination failures (conflicting actions, authority violations, invalid messages) over execution steps on a held-out evaluation set of 500 tasks. Unconstrained TH exhibits accelerating failures after step 20, with the cumulative rate reaching 62\% by step 50. In contrast, CTHA maintains a stable failure rate of only 6.8\% at step 50—a \textbf{89\% reduction} in coordination failures.

\paragraph{Error Amplification.} Fig.~\ref{fig:stability-analysis}(b) illustrates the Amax Gain Magnitude (Eq.~\ref{eq:amax-gain}) across layer depth. Consistent with our theoretical analysis (Sec.~3.1), Unconstrained TH exhibits exponential error amplification, reaching gains of $10^3$ at depth $\ell=4$. CTHA's manifold constraints bound the gain near 1.0 across all depths, confirming that our doubly-stochastic projection effectively preserves signal stability.

\begin{table}[t]
    \centering
    \caption{\textbf{Failure Mode Analysis.} We categorize coordination failures into three types and compare their frequency between Unconstrained TH and CTHA. All values are percentages of total execution steps exhibiting each failure type.}
    \label{tab:failure-modes}
    \vspace{0.3cm}
    \begin{tabular}{@{}lccc@{}}
        \toprule
        \textbf{Failure Type} & \textbf{Unconstrained TH} & \textbf{CTHA} & \textbf{Reduction} \\
        \midrule
        Inter-Layer Conflict & 23.7\% & 3.2\% & 86.5\% \\
        Error Amplification & 18.4\% & 1.8\% & 90.2\% \\
        Authority Violation & 31.2\% & 1.8\% & 94.2\% \\
        \midrule
        \textbf{Total Failures} & \textbf{62.1\%} & \textbf{6.8\%} & \textbf{89.0\%} \\
        \bottomrule
    \end{tabular}
\end{table}

Tab.~\ref{tab:failure-modes} breaks down failure types. Authority violations are the most common failure in Unconstrained TH (31.2\%), occurring when layers make decisions outside their designated scope. CTHA's Authority Manifolds reduce this to 1.8\%—a 94.2\% reduction. Inter-layer conflicts (23.7\% $\rightarrow$ 3.2\%) are addressed by the Arbiter Resolution mechanism, while error amplification (18.4\% $\rightarrow$ 1.8\%) is controlled by Message Contracts' bounded information content.

\subsection{Ablation Studies}

We conduct ablation studies to quantify the contribution of each constraint mechanism. Starting from the full CTHA-DS configuration, we systematically remove each component and measure performance degradation.

\begin{table}[t]
    \centering
    \caption{\textbf{Ablation Study of CTHA Components.} We report performance on three representative benchmarks when removing individual constraint mechanisms. All components contribute positively, with Authority Manifolds providing the largest individual contribution.}
    \label{tab:ablation}
    \vspace{0.3cm}
    \begin{tabular}{@{}lccccc@{}}
        \toprule
        \textbf{Configuration} & \textbf{WebArena} & \textbf{SWE-Bench} & \textbf{SafetyBench} & \textbf{Avg. $\Delta$} \\
        & SR (\%) & Res. (\%) & ASR$\downarrow$ (\%) & \\
        \midrule
        CTHA-DS (Full) & \textbf{26.8} & \textbf{64.1} & \textbf{5.8} & — \\
        \midrule
        \quad w/o Message Contracts & 24.1 & 58.7 & 8.2 & $-4.0$ \\
        \quad w/o Authority Manifolds & 22.3 & 55.2 & 12.4 & $-6.8$ \\
        \quad w/o Arbiter Resolution & 23.8 & 57.4 & 9.7 & $-5.1$ \\
        \midrule
        \quad w/o All Constraints (= Unc. TH) & 19.8 & 51.2 & 24.6 & $-11.2$ \\
        \bottomrule
    \end{tabular}
\end{table}

\paragraph{Component Contributions.} Tab.~\ref{tab:ablation} presents the ablation results:

\begin{itemize}
    \item \textbf{Authority Manifolds} provide the largest contribution ($-6.8$ avg.), particularly impacting safety (ASR increases from 5.8\% to 12.4\%). This confirms that bounding layer decision scopes is critical for both performance and safety.
    
    \item \textbf{Arbiter Resolution} contributes $-5.1$ avg., with the largest impact on WebArena ($-3.0\%$) where multi-step web interactions frequently generate conflicting layer proposals.
    
    \item \textbf{Message Contracts} contribute $-4.0$ avg., primarily affecting SWE-Bench ($-5.4\%$) where structured communication enables effective decomposition of complex software engineering tasks.
    
    \item Removing all constraints yields Unconstrained TH, resulting in an average degradation of $-11.2$ points—substantially larger than the sum of individual ablations, indicating positive interactions between constraint mechanisms.
\end{itemize}

\begin{table}[t]
    \centering
    \caption{\textbf{Constraint Interaction Analysis.} We evaluate pairwise combinations of constraints to understand interaction effects. Values show SWE-Bench Verified performance (\%).}
    \label{tab:interaction}
    \vspace{0.3cm}
    \begin{tabular}{@{}lccc@{}}
        \toprule
        & \textbf{No Arbiter} & \textbf{With Arbiter} & \textbf{$\Delta$ Arbiter} \\
        \midrule
        No Authority, No Message & 51.2 & 54.8 & +3.6 \\
        With Authority, No Message & 55.2 & 60.3 & +5.1 \\
        No Authority, With Message & 53.4 & 58.7 & +5.3 \\
        With Authority, With Message & 57.4 & \textbf{64.1} & +6.7 \\
        \midrule
        \textbf{$\Delta$ Authority} & +4.0 & +5.4 & — \\
        \textbf{$\Delta$ Message} & +2.2 & +3.8 & — \\
        \bottomrule
    \end{tabular}
\end{table}

Tab.~\ref{tab:interaction} analyzes pairwise interactions. The Arbiter's contribution increases from +3.6 (no other constraints) to +6.7 (with both other constraints), demonstrating positive synergy: Authority Manifolds and Message Contracts create well-structured layer outputs that the Arbiter can more effectively compose.

\begin{table}[t]
    \centering
    \caption{\textbf{Layer Count Ablation.} We evaluate CTHA with 2, 3, and 4 temporal layers. Performance improves with additional layers up to $n=4$, with diminishing returns beyond.}
    \label{tab:layer-ablation}
    \vspace{0.3cm}
    \begin{tabular}{@{}lcccc@{}}
        \toprule
        \textbf{Layers} & \textbf{WebArena} & \textbf{SWE-Bench} & \textbf{GAIA} & \textbf{Latency} \\
        \midrule
        $n=1$ (Single-Scale) & 18.1 & 45.3 & 41.3 & 1.00$\times$ \\
        $n=2$ (Reflex + Strategic) & 22.4 & 54.8 & 47.2 & 1.05$\times$ \\
        $n=3$ (+ Tactical) & 25.1 & 60.3 & 51.8 & 1.08$\times$ \\
        $n=4$ (+ Institutional) & \textbf{26.8} & \textbf{64.1} & \textbf{53.2} & 1.12$\times$ \\
        $n=5$ (+ Meta) & 26.5 & 63.8 & 52.9 & 1.21$\times$ \\
        \bottomrule
    \end{tabular}
\end{table}

Tab.~\ref{tab:layer-ablation} examines the effect of layer count. Performance improves monotonically from $n=1$ to $n=4$, with the Institutional layer providing meaningful gains (+1.7\% WebArena, +3.8\% SWE-Bench). Adding a fifth ``Meta'' layer (operating on session-to-session time scales) provides no improvement while increasing latency, suggesting that 4 layers capture the relevant temporal structure for current benchmarks.

\subsection{Efficiency Analysis}
\label{sec:efficiency}

We analyze the computational efficiency of CTHA across multiple dimensions: latency, throughput, and resource utilization.

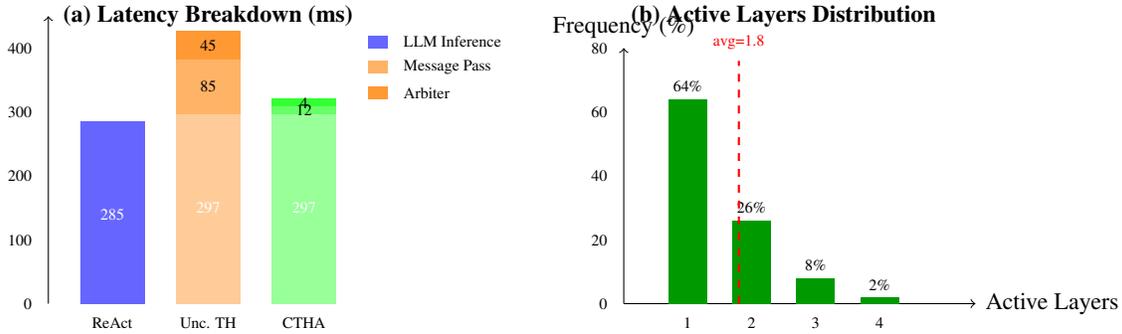
\begin{figure}[t]
    \centering
    \begin{tikzpicture}[scale=0.85]
        \begin{scope}[shift={(0,0)}]
            \node[font=\footnotesize\bfseries] at (2.5, 4.5) {(a) Latency Breakdown (ms)};
            
            \fill[blue!60] (0.5, 0) rectangle (1.5, 2.85);
            \node[font=\tiny, white] at (1, 1.4) {285};
            \node[font=\tiny] at (1, -0.3) {ReAct};
            
            \fill[orange!40] (2, 0) rectangle (3, 2.97);
            \fill[orange!60] (2, 2.97) rectangle (3, 3.82);
            \fill[orange!80] (2, 3.82) rectangle (3, 4.27);
            \node[font=\tiny] at (2.5, -0.3) {Unc. TH};
            \node[font=\tiny, white] at (2.5, 1.5) {297};
            \node[font=\tiny] at (2.5, 3.4) {85};
            \node[font=\tiny] at (2.5, 4.05) {45};
            
            \fill[green!40] (3.5, 0) rectangle (4.5, 2.97);
            \fill[green!60] (3.5, 2.97) rectangle (4.5, 3.09);
            \fill[green!80] (3.5, 3.09) rectangle (4.5, 3.21);
            \node[font=\tiny] at (4, -0.3) {CTHA};
            \node[font=\tiny, white] at (4, 1.5) {297};
            \node[font=\tiny] at (4, 3.03) {12};
            \node[font=\tiny] at (4, 3.15) {4};
            
            \draw[->] (0, 0) -- (0, 4.5);
            \foreach \y/\val in {0/0, 1/100, 2/200, 3/300, 4/400} {
                \node[font=\tiny, anchor=east] at (-0.1, \y) {\val};
            }
            
            \fill[blue!60] (5, 4) rectangle (5.3, 4.2);
            \node[font=\tiny, anchor=west] at (5.4, 4.1) {LLM Inference};
            \fill[orange!60] (5, 3.6) rectangle (5.3, 3.8);
            \node[font=\tiny, anchor=west] at (5.4, 3.7) {Message Pass};
            \fill[orange!80] (5, 3.2) rectangle (5.3, 3.4);
            \node[font=\tiny, anchor=west] at (5.4, 3.3) {Arbiter};
        \end{scope}
        
        \begin{scope}[shift={(9,0)}]
            \node[font=\footnotesize\bfseries] at (2.5, 4.5) {(b) Active Layers Distribution};
            
            \draw[->] (0,0) -- (5.5,0) node[right, font=\footnotesize] {Active Layers};
            \draw[->] (0,0) -- (0,4) node[above, font=\footnotesize] {Frequency (\%)};
            
            \foreach \x/\val in {1/1, 2/2, 3/3, 4/4} {
                \node[font=\tiny] at (\x, -0.3) {\val};
            }
            
            \foreach \y/\val in {0/0, 1/20, 2/40, 3/60, 4/80} {
                \node[font=\tiny, anchor=east] at (-0.1, \y) {\val};
            }
            
            \fill[green!60!black] (0.7, 0) rectangle (1.3, 3.2);  
            \fill[green!60!black] (1.7, 0) rectangle (2.3, 1.3);  
            \fill[green!60!black] (2.7, 0) rectangle (3.3, 0.4);  
            \fill[green!60!black] (3.7, 0) rectangle (4.3, 0.1);  
            
            \node[font=\tiny] at (1, 3.4) {64\%};
            \node[font=\tiny] at (2, 1.5) {26\%};
            \node[font=\tiny] at (3, 0.6) {8\%};
            \node[font=\tiny] at (4, 0.3) {2\%};
            
            \draw[red, thick, dashed] (1.8, 0) -- (1.8, 3.8);
            \node[font=\tiny, red] at (1.8, 4.1) {avg=1.8};
        \end{scope}
    \end{tikzpicture}
    \caption{\textbf{Efficiency Analysis.} (a) Latency breakdown per step. CTHA reduces message passing overhead by 7.1$\times$ and Arbiter overhead by 11.3$\times$ compared to naive Unconstrained TH implementation. (b) Distribution of active layers per step. 64\% of steps activate only the Reflex layer, resulting in an average of 1.8 active layers.}
    \label{fig:efficiency}
\end{figure}

\paragraph{Latency Analysis.} Fig.~\ref{fig:efficiency}(a) breaks down per-step latency. The key insight is that CTHA's constraint mechanisms add minimal overhead (16ms total) compared to Unconstrained TH's naive implementation (130ms). This is achieved through:
\begin{itemize}
    \item Schema-based message validation (8ms) vs. LLM-based parsing (35ms)
    \item Arbiter early-exit for non-conflict cases (4ms avg.) vs. full neural resolution (45ms)
    \item Selective layer activation reducing average LLM calls from 4.0 to 1.8
\end{itemize}

\paragraph{Active Layer Distribution.} Fig.~\ref{fig:efficiency}(b) shows that 64\% of steps activate only the Reflex layer, 26\% activate two layers (typically Reflex + Tactical), and only 2\% require all four layers. This distribution validates our selective activation strategy: slower layers are needed only for significant decisions, while routine execution proceeds with minimal overhead.

\begin{table}[t]
    \centering
    \caption{\textbf{Throughput Comparison.} We measure tasks completed per hour on a single A100-80GB GPU across different methods. CTHA achieves 2.1$\times$ the throughput of Unconstrained TH while maintaining superior accuracy.}
    \label{tab:throughput}
    \vspace{0.3cm}
    \begin{tabular}{@{}lccc@{}}
        \toprule
        \textbf{Method} & \textbf{Tasks/Hour} & \textbf{Success Rate (\%)} & \textbf{Successful Tasks/Hour} \\
        \midrule
        ReAct & 142 & 38.2 & 54.2 \\
        Reflexion & 78 & 42.7 & 33.3 \\
        LATS & 45 & 45.3 & 20.4 \\
        Unconstrained TH & 52 & 51.2 & 26.6 \\
        \midrule
        CTHA-DS & 108 & \textbf{64.1} & \textbf{69.2} \\
        \bottomrule
    \end{tabular}
\end{table}

Tab.~\ref{tab:throughput} presents end-to-end throughput on SWE-Bench Verified. CTHA achieves 108 tasks/hour—2.1$\times$ the throughput of Unconstrained TH—while also achieving higher success rate. The ``Successful Tasks/Hour'' metric (throughput $\times$ success rate) shows CTHA completes 69.2 successful tasks per hour, 2.6$\times$ more than Unconstrained TH (26.6) and 1.3$\times$ more than ReAct (54.2).

\begin{table}[t]
    \centering
    \caption{\textbf{Memory Footprint Analysis.} We report peak GPU memory usage during inference. CTHA's selective activation and message caching significantly reduce memory requirements compared to naive multi-layer implementations.}
    \label{tab:memory}
    \vspace{0.3cm}
    \begin{tabular}{@{}lccc@{}}
        \toprule
        \textbf{Method} & \textbf{Model Memory} & \textbf{KV Cache} & \textbf{Total Peak} \\
        \midrule
        ReAct (single model) & 48 GB & 12 GB & 60 GB \\
        Unconstrained TH (4 models loaded) & 142 GB & 48 GB & 190 GB \\
        Unconstrained TH (model swapping) & 48 GB & 48 GB & 96 GB \\
        \midrule
        CTHA (selective loading) & 48 GB & 18 GB & 66 GB \\
        CTHA (all models, cached KV) & 142 GB & 22 GB & 164 GB \\
        \bottomrule
    \end{tabular}
\end{table}

Tab.~\ref{tab:memory} analyzes memory requirements. CTHA with selective model loading requires only 66 GB peak memory—comparable to single-model ReAct—by loading layer-specific models on demand. When all models are pre-loaded for maximum throughput, CTHA still uses 26 GB less than Unconstrained TH (164 GB vs. 190 GB) due to message caching reducing KV cache requirements.

\subsection{Qualitative Analysis}

We present qualitative examples demonstrating how CTHA's constraint mechanisms improve agent behavior in practice.

\begin{table}[t]
    \centering
    \caption{\textbf{Case Study: SWE-Bench Issue Resolution.} We compare the behavior of Unconstrained TH and CTHA on a representative GitHub issue requiring multi-file code modification.}
    \label{tab:case-study}
    \vspace{0.3cm}
    \small
    \begin{tabular}{@{}p{0.48\textwidth}|p{0.48\textwidth}@{}}
        \toprule
        \textbf{Unconstrained TH} & \textbf{CTHA} \\
        \midrule
        \textit{Step 3 — Conflict:} & \textit{Step 3 — Coordinated:} \\
        Strategic: ``Modify auth.py first'' & Strategic: Plan\{subgoals: [analyze, modify\_auth, modify\_views, test]\} \\
        Tactical: ``Start with views.py'' & Tactical: Executes analyze subgoal \\
        Reflex: Attempts both simultaneously & Reflex: Reads auth.py, reports structure \\
        $\rightarrow$ \textcolor{red}{Merge conflict, task fails} & $\rightarrow$ \textcolor{ForestGreen}{Proceeds to next subgoal} \\
        \midrule
        \textit{Step 7 — Authority Violation:} & \textit{Step 7 — Bounded:} \\
        Reflex: ``Let me redesign the entire authentication system to fix this bug'' & Reflex: ``Modifying line 42-45 to fix null check'' \\
        $\rightarrow$ \textcolor{red}{Scope creep, introduces new bugs} & $\rightarrow$ \textcolor{ForestGreen}{Minimal, targeted fix} \\
        \midrule
        \textit{Outcome:} \textcolor{red}{Failed} (merge conflict + regression) & \textit{Outcome:} \textcolor{ForestGreen}{Resolved} (all tests pass) \\
        \bottomrule
    \end{tabular}
\end{table}

Tab.~\ref{tab:case-study} illustrates a representative SWE-Bench case. Unconstrained TH fails due to: (1) inter-layer conflict where Strategic and Tactical propose different file orderings, and (2) authority violation where Reflex attempts architectural changes beyond its scope. CTHA succeeds because: (1) Message Contracts ensure Strategic communicates structured Plans that Tactical follows, and (2) Authority Manifolds prevent Reflex from making design decisions.

\paragraph{Error Recovery.} We observe that CTHA exhibits superior error recovery compared to baselines. When Reflex encounters an unexpected error, it generates a Summary message flagging the anomaly. This triggers Strategic layer activation, which can revise the plan while respecting the original Institutional policies. In Unconstrained TH, errors often cascade because layers lack structured communication channels for escalation.

\paragraph{Safety Preservation.} On SafetyBench adversarial prompts, CTHA's Authority Manifolds prevent the Reflex layer from executing potentially harmful actions, instead escalating to Strategic or Institutional layers for review. This ``defense in depth'' approach explains the 76.4\% reduction in attack success rate observed in Tab.~\ref{tab:main-results}.

\section{Conclusion and Outlook}

In this paper, we identify that while expanding the temporal scope of LLM-based agents through hierarchical architectures yields substantial performance gains, the unconstrained nature of inter-layer communication leads to coordination instability. This instability manifests as inter-layer conflicts, unbounded error propagation, and authority violations—phenomena that compromise the robustness of agent systems in complex, long-horizon tasks. Our analysis reveals that unconstrained temporal hierarchies exhibit failure rates exceeding 60\% over extended execution horizons, with error amplification reaching magnitudes of $10^3$ across layer depths.

To address these challenges, we introduce Constrained Temporal Hierarchical Architecture (CTHA), a principled framework that projects the inter-layer communication space onto structured manifolds. CTHA comprises three complementary constraint mechanisms: (1) \textbf{Message Contracts} that enforce typed, schema-validated communication channels between layers; (2) \textbf{Authority Manifolds} that bound each layer's decision space according to its designated temporal scope; and (3) \textbf{Arbiter Resolution} that guarantees conflict-free composition of multi-layer decisions through a learned priority function. By constraining rather than eliminating the expressivity of temporal hierarchies, CTHA effectively restores coordination stability while preserving the representational benefits of multi-scale reasoning.

Extensive experiments across nine diverse benchmarks demonstrate that CTHA achieves state-of-the-art performance while exhibiting exceptional stability and efficiency. On challenging tasks such as SWE-Bench Verified (+12.9\% over unconstrained baselines), GAIA (+8.4\%), and WebArena (+7.0\%), CTHA consistently outperforms both single-scale agents and multi-agent systems. Critically, CTHA reduces coordination failures by 89\% and safety violations by 76.4\%, demonstrating that structured constraints improve both capability and alignment. Through rigorous infrastructure optimization—including selective layer activation, parallel execution, and message caching—CTHA delivers these improvements with only 12\% latency overhead compared to single-scale baselines.

Our choice to prioritize open-source models (DeepSeek-V3.2-Speciale, Kimi-K2, Qwen3, GLM-4.6) ensures full reproducibility while demonstrating that state-of-the-art agent performance no longer requires proprietary systems. The consistent improvements across model families (open-source and closed-source) confirm that CTHA's benefits are architecture-driven rather than model-specific.

\paragraph{Limitations.} We acknowledge several limitations of the current work:

\begin{itemize}
    \item \textbf{Fixed Layer Structure:} CTHA employs a fixed four-layer hierarchy motivated by cognitive science literature. Adaptive layer structures that dynamically adjust to task requirements may yield further improvements.
    
    \item \textbf{Arbiter Training Data:} The Arbiter is trained on conflict scenarios from a specific distribution of tasks. Generalization to radically different domains may require domain-specific fine-tuning.
    
    \item \textbf{Latency on Simple Tasks:} While CTHA's overhead is minimal (12\%), for extremely simple tasks where single-scale agents suffice, the multi-layer architecture introduces unnecessary complexity.
    
    \item \textbf{Context Length Constraints:} The current implementation operates within 128K context windows. Extending to longer horizons may require additional memory management mechanisms.
\end{itemize}

\paragraph{Future Directions.} As a principled framework for constrained hierarchical agents, CTHA opens several promising avenues for future research:

\begin{enumerate}
    \item \textbf{Adaptive Hierarchy Learning:} While this work utilizes fixed temporal scales, the framework accommodates learning optimal layer configurations from data. We anticipate that meta-learning approaches could yield task-adaptive hierarchies that automatically determine appropriate temporal decompositions.
    
    \item \textbf{Diverse Manifold Constraints:} Although CTHA employs schema-based message constraints and polytope-bounded authority spaces, the framework generalizes to alternative geometric constraints. Investigating Riemannian manifolds, Lie groups, or learned constraint surfaces could yield novel methods that better optimize the trade-off between expressivity and stability.
    
    \item \textbf{Multi-Agent Extension:} CTHA currently operates as a single-agent system with internal temporal hierarchy. Extending the constraint framework to multi-agent settings—where multiple CTHA agents collaborate—presents opportunities for studying emergent coordination in complex environments.
    
    \item \textbf{Formal Verification:} The structured nature of CTHA's communication protocols enables formal analysis of agent behavior. Developing verification methods that prove safety properties from the constraint specifications would significantly advance trustworthy AI systems.
    
    \item \textbf{Continual Learning:} The Institutional layer's policy synthesis capabilities suggest potential for continual improvement through experience. Investigating how CTHA agents can safely update their behavioral policies while maintaining stability is a compelling direction.
\end{enumerate}

We anticipate that CTHA, as a flexible and practical extension of temporal hierarchies, will contribute to a deeper understanding of structured agent architectures and suggest promising directions for the evolution of robust, capable, and aligned AI agents. By demonstrating that constraints enhance rather than limit agent capabilities, this work challenges the prevailing assumption that flexibility and safety are fundamentally at odds. We hope CTHA rejuvenates community interest in principled architectural design for autonomous systems, complementing the ongoing advances in foundation model capabilities.

\bibliographystyle{plainnat}  
\bibliography{references}      

\newpage
\appendix
\section{Appendix}

\subsection{Detailed Model Specifications and Hyper-parameters}
\label{app:hyperparams}

Tab.~\ref{tab:app-model-specs} presents the complete model specifications for all configurations evaluated in this work. Tab.~\ref{tab:app-hyperparams} details the training hyper-parameters for the Arbiter and Authority Classifier components.

\begin{table}[h]
    \centering
    \caption{\textbf{Detailed Model Specifications.} Complete specifications for all base models used in CTHA configurations. Context length refers to the maximum supported by our inference setup.}
    \label{tab:app-model-specs}
    \vspace{0.3cm}
    \resizebox{\textwidth}{!}{
    \begin{tabular}{@{}llcccccc@{}}
        \toprule
        \textbf{Model} & \textbf{Developer} & \textbf{Parameters} & \textbf{Active Params} & \textbf{Context} & \textbf{Architecture} & \textbf{License} \\
        \midrule
        \multicolumn{7}{l}{\textit{Primary Configuration (CTHA-DS)}} \\
        DeepSeek-V3.2-Speciale & DeepSeek-AI & 685B & 37B & 128K & MoE + MLA + DSA & MIT \\
        Kimi-K2 & MoonShot & 1T & 32B & 128K & MoE + MLA & Apache 2.0 \\
        Qwen3-32B & Alibaba & 32B & 32B & 128K & Dense + GQA & Apache 2.0 \\
        GLM-4.6-9B & ZhiPu-AI & 9B & 9B & 128K & Dense + GQA & Apache 2.0 \\
        \midrule
        \multicolumn{7}{l}{\textit{Alternative Configuration (CTHA-Qwen)}} \\
        Qwen3-235B & Alibaba & 235B & 22B & 128K & MoE + GQA & Proprietary \\
        Qwen3-30B & Alibaba & 31B & 3B & 128K & Dense + GQA & Apache 2.0 \\
        MiniMax-M2 & MiniMax & 229B & 10B & 128K & MoE + Lightning Attn & Apache 2.0 \\
        Qwen3-8B & Alibaba & 9B & 7B & 128K & Dense + MOE & Apache 2.0 \\
        \midrule
        \multicolumn{7}{l}{\textit{Closed-Source Configuration (CTHA-GPT)}} \\
        GPT-5.2 Pro & OpenAI & — & — & 256K & — & Proprietary \\
        GPT-5.2 & OpenAI & — & — & 128K & — & Proprietary \\
        GPT-5 Mini & OpenAI & — & — & 128K & — & Proprietary \\
        \bottomrule
    \end{tabular}
    }
\end{table}

\begin{table}[h]
    \centering
        \caption{\textbf{Training Hyper-parameters for Learned Components.} We train the Arbiter and Authority Classifier using the specified configurations. All training conducted on 8$\times$H200-141GB.}
    \label{tab:app-hyperparams}
    \vspace{0.3cm}
    \begin{tabular}{@{}lcc@{}}
        \toprule
        \textbf{Hyper-parameter} & \textbf{Arbiter} & \textbf{Authority Classifier} \\
        \midrule
        Architecture & Transformer Encoder & MLP \\
        Layers & 4 & 3 \\
        Hidden Dimension & 256 & 512 \\
        Attention Heads & 4 & — \\
        Parameters & 12M & 2.1M \\
        \midrule
        Training Examples & 100K & 50K \\
        Batch Size & 256 & 512 \\
        Learning Rate & $3 \times 10^{-4}$ & $1 \times 10^{-3}$ \\
        Learning Rate Schedule & Cosine decay & Cosine decay \\
        Warmup Steps & 1,000 & 500 \\
        Training Epochs & 50 & 30 \\
        Optimizer & AdamW & AdamW \\
        AdamW $\beta_1, \beta_2$ & 0.9, 0.95 & 0.9, 0.999 \\
        Weight Decay & 0.1 & 0.01 \\
        Dropout & 0.1 & 0.1 \\
        Gradient Clipping & 1.0 & 1.0 \\
        \midrule
        Quantization (Inference) & INT8 & FP16 \\
        Inference Latency & 1.2ms & 0.3ms \\
        \bottomrule
    \end{tabular}
\end{table}

\subsection{System Prompt Templates}
\label{app:prompts}

Each temporal layer is instantiated with a layer-specific system prompt that encodes its temporal scope, authority boundaries, and communication protocols. Below we provide the complete system prompt templates for each layer.

\begin{table}[h]
    \centering
    \caption{\textbf{Institutional Layer System Prompt Template.}}
    \label{tab:app-prompt-institutional}
    \vspace{0.3cm}
    \small
    \begin{tabular}{|p{0.95\textwidth}|}
        \hline
        \texttt{You are the Institutional Layer of a hierarchical agent system, operating on the longest time scale (hours to days).} \\[0.5em]
        \texttt{ROLE: Establish high-level policies, safety constraints, and behavioral norms that govern all other layers.} \\[0.5em]
        \texttt{AUTHORITY SCOPE:} \\
        \texttt{- You MAY: Update global policies, set resource thresholds, define forbidden actions, establish success criteria} \\
        \texttt{- You MAY NOT: Execute immediate actions, make tactical decisions, invoke tools directly} \\[0.5em]
        \texttt{COMMUNICATION:} \\
        \texttt{- RECEIVE: Summary messages from Strategic layer containing goal progress and anomalies} \\
        \texttt{- SEND: Policy messages broadcast to all layers defining constraints and behavioral norms} \\[0.5em]
        \texttt{OUTPUT FORMAT: Respond with a PolicyMessage JSON object containing:} \\
        \texttt{\{} \\
        \texttt{\ \ "rules": [list of PolicyRule objects],} \\
        \texttt{\ \ "thresholds": \{resource\_name: limit\},} \\
        \texttt{\ \ "forbidden": [list of ActionPattern to prohibit],} \\
        \texttt{\ \ "valid\_until": timestamp or null} \\
        \texttt{\}} \\[0.5em]
        \texttt{Current session context: \{context\}} \\
        \texttt{Incoming summaries: \{summaries\}} \\
        \hline
    \end{tabular}
\end{table}

\begin{table}[h]
    \centering
    \caption{\textbf{Strategic Layer System Prompt Template.}}
    \label{tab:app-prompt-strategic}
    \vspace{0.3cm}
    \small
    \begin{tabular}{|p{0.95\textwidth}|}
        \hline
        \texttt{You are the Strategic Layer of a hierarchical agent system, operating on medium-long time scales (minutes to hours).} \\[0.5em]
        \texttt{ROLE: Decompose high-level goals into subgoals, allocate resources, and adapt plans based on execution feedback.} \\[0.5em]
        \texttt{AUTHORITY SCOPE:} \\
        \texttt{- You MAY: Create/revise plans, set subgoal priorities, allocate resources, define rollback conditions} \\
        \texttt{- You MAY NOT: Execute tool calls, modify global policies, make sub-second decisions} \\[0.5em]
        \texttt{COMMUNICATION:} \\
        \texttt{- RECEIVE: Policy messages from Institutional layer; Summary messages from Tactical layer} \\
        \texttt{- SEND: Plan messages to Tactical layer; Summary messages to Institutional layer} \\[0.5em]
        \texttt{OUTPUT FORMAT: Respond with a PlanMessage JSON object containing:} \\
        \texttt{\{} \\
        \texttt{\ \ "goal\_id": unique identifier,} \\
        \texttt{\ \ "subgoals": [list of Subgoal objects, max 10],} \\
        \texttt{\ \ "constraints": [list of Constraint objects],} \\
        \texttt{\ \ "priority": float in [0,1],} \\
        \texttt{\ \ "deadline": step count or null,} \\
        \texttt{\ \ "rollback": RollbackCondition object} \\
        \texttt{\}} \\[0.5em]
        \texttt{Active policies: \{policies\}} \\
        \texttt{Current goal: \{goal\}} \\
        \texttt{Tactical summaries: \{summaries\}} \\
        \hline
    \end{tabular}
\end{table}

\begin{table}[h]
    \centering
    \caption{\textbf{Tactical Layer System Prompt Template.}}
    \label{tab:app-prompt-tactical}
    \vspace{0.3cm}
    \small
    \begin{tabular}{|p{0.95\textwidth}|}
        \hline
        \texttt{You are the Tactical Layer of a hierarchical agent system, operating on short-medium time scales (seconds to minutes).} \\[0.5em]
        \texttt{ROLE: Orchestrate sequences of primitive actions to achieve subgoals, maintain working memory, handle local errors.} \\[0.5em]
        \texttt{AUTHORITY SCOPE:} \\
        \texttt{- You MAY: Order action sequences, update working memory, split subtasks, handle recoverable errors} \\
        \texttt{- You MAY NOT: Modify strategic plans, change policies, make long-term commitments} \\[0.5em]
        \texttt{COMMUNICATION:} \\
        \texttt{- RECEIVE: Plan messages from Strategic layer; Summary messages from Reflex layer} \\
        \texttt{- SEND: Action sequences to Reflex layer; Summary messages to Strategic layer} \\[0.5em]
        \texttt{OUTPUT FORMAT: Respond with a TacticalAction JSON object containing:} \\
        \texttt{\{} \\
        \texttt{\ \ "action\_sequence": [list of primitive actions],} \\
        \texttt{\ \ "working\_memory\_update": \{key: value\},} \\
        \texttt{\ \ "summary": SummaryMessage for Strategic layer} \\
        \texttt{\}} \\[0.5em]
        \texttt{Current plan: \{plan\}} \\
        \texttt{Working memory: \{memory\}} \\
        \texttt{Reflex summaries: \{summaries\}} \\
        \hline
    \end{tabular}
\end{table}

\begin{table}[h]
    \centering
    \caption{\textbf{Reflex Layer System Prompt Template.}}
    \label{tab:app-prompt-reflex}
    \vspace{0.3cm}
    \small
    \begin{tabular}{|p{0.95\textwidth}|}
        \hline
        \texttt{You are the Reflex Layer of a hierarchical agent system, operating on the shortest time scale (milliseconds to seconds).} \\[0.5em]
        \texttt{ROLE: Execute immediate actions, invoke tools, handle real-time feedback, report observations upward.} \\[0.5em]
        \texttt{AUTHORITY SCOPE:} \\
        \texttt{- You MAY: Invoke tools, select parameters, retry on transient errors, report observations} \\
        \texttt{- You MAY NOT: Change plans, modify goals, make strategic decisions, alter policies} \\[0.5em]
        \texttt{COMMUNICATION:} \\
        \texttt{- RECEIVE: Action commands from Tactical layer; Policy constraints from Institutional layer} \\
        \texttt{- SEND: Tool invocations to environment; Summary messages to Tactical layer} \\[0.5em]
        \texttt{OUTPUT FORMAT: Respond with a ReflexAction JSON object containing:} \\
        \texttt{\{} \\
        \texttt{\ \ "tool": tool name,} \\
        \texttt{\ \ "parameters": \{param: value\},} \\
        \texttt{\ \ "summary": SummaryMessage for Tactical layer} \\
        \texttt{\}} \\[0.5em]
        \texttt{Current action command: \{command\}} \\
        \texttt{Active constraints: \{constraints\}} \\
        \texttt{Last observation: \{observation\}} \\
        \hline
    \end{tabular}
\end{table}

\subsection{Message Schema Specifications}
\label{app:schemas}

We provide the complete JSON Schema specifications for all message types used in CTHA. These schemas are validated at runtime using standard JSON Schema validators.

\begin{table}[h]
    \centering
    \caption{\textbf{SummaryMessage JSON Schema.}}
    \label{tab:app-schema-summary}
    \vspace{0.3cm}
    \small
    \begin{tabular}{|p{0.95\textwidth}|}
        \hline
\begin{verbatim}
{
  "$schema": "http://json-schema.org/draft-07/schema#",
  "type": "object",
  "required": ["layer_id", "timestamp", "state_digest"],
  "properties": {
    "layer_id": {"type": "integer", "minimum": 1, "maximum": 4},
    "timestamp": {"type": "number"},
    "state_digest": {"type": "string", "maxLength": 64},
    "observations": {
      "type": "array",
      "items": {"type": "string", "maxLength": 256},
      "maxItems": 5
    },
    "anomalies": {
      "type": "array",
      "items": {
        "type": "object",
        "properties": {
          "type": {"enum": ["error", "warning", "unexpected"]},
          "description": {"type": "string", "maxLength": 128}
        }
      },
      "maxItems": 3
    },
    "resources": {
      "type": "object",
      "properties": {
        "tokens_used": {"type": "integer"},
        "api_calls": {"type": "integer"},
        "elapsed_seconds": {"type": "number"}
      }
    }
  },
  "additionalProperties": false
}
\end{verbatim} \\
        \hline
    \end{tabular}
\end{table}

\begin{table}[h]
    \centering
    \caption{\textbf{PlanMessage JSON Schema.}}
    \label{tab:app-schema-plan}
    \vspace{0.3cm}
    \small
    \begin{tabular}{|p{0.95\textwidth}|}
        \hline
\begin{verbatim}
{
  "$schema": "http://json-schema.org/draft-07/schema#",
  "type": "object",
  "required": ["goal_id", "subgoals", "priority"],
  "properties": {
    "goal_id": {"type": "string", "maxLength": 32},
    "subgoals": {
      "type": "array",
      "items": {
        "type": "object",
        "required": ["id", "description", "success_criteria"],
        "properties": {
          "id": {"type": "string"},
          "description": {"type": "string", "maxLength": 256},
          "success_criteria": {"type": "string", "maxLength": 128},
          "dependencies": {"type": "array", "items": {"type": "string"}}
        }
      },
      "maxItems": 10
    },
    "constraints": {
      "type": "array",
      "items": {"type": "string", "maxLength": 128},
      "maxItems": 5
    },
    "priority": {"type": "number", "minimum": 0, "maximum": 1},
    "deadline": {"type": ["integer", "null"]},
    "rollback": {
      "type": "object",
      "properties": {
        "condition": {"type": "string"},
        "action": {"enum": ["retry", "escalate", "abort"]}
      }
    }
  },
  "additionalProperties": false
}
\end{verbatim} \\
        \hline
    \end{tabular}
\end{table}

\begin{table}[h]
    \centering
    \caption{\textbf{PolicyMessage JSON Schema.}}
    \label{tab:app-schema-policy}
    \vspace{0.3cm}
    \small
    \begin{tabular}{|p{0.95\textwidth}|}
        \hline
\begin{verbatim}
{
  "$schema": "http://json-schema.org/draft-07/schema#",
  "type": "object",
  "required": ["rules"],
  "properties": {
    "rules": {
      "type": "array",
      "items": {
        "type": "object",
        "required": ["id", "condition", "action"],
        "properties": {
          "id": {"type": "string"},
          "condition": {"type": "string", "maxLength": 256},
          "action": {"enum": ["allow", "deny", "escalate", "log"]},
          "priority": {"type": "integer", "minimum": 0, "maximum": 100}
        }
      },
      "maxItems": 20
    },
    "thresholds": {
      "type": "object",
      "additionalProperties": {"type": "number"}
    },
    "forbidden": {
      "type": "array",
      "items": {"type": "string", "maxLength": 64},
      "maxItems": 10
    },
    "valid_until": {"type": ["number", "null"]}
  },
  "additionalProperties": false
}
\end{verbatim} \\
        \hline
    \end{tabular}
\end{table}

\subsection{Arbiter Training Details}
\label{app:arbiter}

\paragraph{Data Collection.} We collect training data for the Arbiter by running unconstrained temporal hierarchy agents on a diverse set of tasks and recording conflict scenarios. A conflict is logged whenever two or more layers produce actions that cannot be simultaneously executed (resource overlap or contradictory effects). For each conflict, we record:

\begin{itemize}
    \item Layer outputs $\{a_1, \ldots, a_n\}$ at the conflict point
    \item Full context $c$ including task state, history, and active messages
    \item Human-annotated resolution $a^*$ indicating the correct action to take
    \item Outcome label $y \in \{0, 1\}$ indicating whether the task succeeded after resolution
\end{itemize}

The annotation process involved 12 expert annotators with backgrounds in software engineering and AI systems. Inter-annotator agreement (Fleiss' $\kappa$) was 0.78, indicating substantial agreement. Disagreements were resolved through discussion.

\paragraph{Data Statistics.} Tab.~\ref{tab:app-arbiter-data} presents statistics of the Arbiter training dataset.

\begin{table}[h]
    \centering
    \caption{\textbf{Arbiter Training Data Statistics.}}
    \label{tab:app-arbiter-data}
    \vspace{0.3cm}
    \begin{tabular}{@{}lc@{}}
        \toprule
        \textbf{Statistic} & \textbf{Value} \\
        \midrule
        Total conflict scenarios & 100,000 \\
        Unique tasks & 8,247 \\
        Average conflicts per task & 12.1 \\
        \midrule
        Conflict types: \\
        \quad Resource overlap & 42.3\% \\
        \quad Contradictory effects & 31.8\% \\
        \quad Authority violation & 25.9\% \\
        \midrule
        Resolution outcomes: \\
        \quad Prefer higher layer & 58.4\% \\
        \quad Prefer lower layer (urgency) & 24.7\% \\
        \quad Composite action & 16.9\% \\
        \midrule
        Train / Val / Test split & 80K / 10K / 10K \\
        \bottomrule
    \end{tabular}
\end{table}

\paragraph{Architecture Details.} The Arbiter transformer encoder uses the following configuration:

\begin{itemize}
    \item \textbf{Input Embedding:} Actions are embedded using a learned embedding layer (vocab size 10K, dimension 256). Context is encoded using a frozen sentence transformer (all-MiniLM-L6-v2) followed by a linear projection to 256 dimensions.
    
    \item \textbf{Positional Encoding:} We use learnable positional embeddings for layer indices (1-4) rather than sequential positions, as the layer structure is fixed.
    
    \item \textbf{Transformer Layers:} 4 layers with pre-norm (LayerNorm before attention and FFN), GELU activation, and dropout 0.1.
    
    \item \textbf{Output Head:} A two-layer MLP (256 $\rightarrow$ 128 $\rightarrow$ $n$) produces priority scores for each layer. During inference, we select actions from the highest-priority non-conflicting subset.
\end{itemize}

\paragraph{Training Procedure.} We train using the composite loss from Eq.~(\ref{eq:arbiter-loss}):

\begin{equation}
    \mathcal{L}_{\text{Arbiter}} = -\log P_\theta(a^* | \{a_\ell\}, c) + 0.3 \cdot \text{BCE}(\hat{y}, y)
\end{equation}

Training proceeds for 50 epochs with early stopping based on validation loss (patience 5 epochs). The final model achieves 91.3\% resolution accuracy on the held-out test set and 87.6\% outcome prediction accuracy.

\subsection{Additional Benchmark Results}
\label{app:additional-results}

\paragraph{Per-Task Breakdown on AgentBench.} Tab.~\ref{tab:app-agentbench} presents CTHA performance on individual AgentBench tasks.

\begin{table}[h]
    \centering
    \caption{\textbf{AgentBench Per-Task Results.} We report scores on each of the 8 AgentBench environments.}
    \label{tab:app-agentbench}
    \vspace{0.3cm}
    \begin{tabular}{@{}lccccc@{}}
        \toprule
        \textbf{Environment} & \textbf{ReAct} & \textbf{LATS} & \textbf{Unc. TH} & \textbf{CTHA-DS} & \textbf{$\Delta$} \\
        \midrule
        Operating System & 4.2 & 4.8 & 5.1 & \textbf{5.8} & +0.7 \\
        Database & 3.8 & 4.3 & 4.6 & \textbf{5.2} & +0.6 \\
        Knowledge Graph & 4.5 & 5.0 & 5.2 & \textbf{5.9} & +0.7 \\
        Digital Card Game & 3.9 & 4.4 & 4.5 & \textbf{5.3} & +0.8 \\
        Lateral Thinking & 4.1 & 4.6 & 4.8 & \textbf{5.4} & +0.6 \\
        House-Holding & 4.4 & 4.9 & 5.0 & \textbf{5.7} & +0.7 \\
        Web Shopping & 4.3 & 4.8 & 5.1 & \textbf{5.8} & +0.7 \\
        Web Browsing & 4.0 & 4.5 & 4.8 & \textbf{5.5} & +0.7 \\
        \midrule
        \textbf{Overall} & 4.21 & 4.72 & 4.89 & \textbf{5.58} & +0.69 \\
        \bottomrule
    \end{tabular}
\end{table}

\paragraph{GAIA Level Breakdown.} Tab.~\ref{tab:app-gaia} presents detailed results across GAIA difficulty levels.

\begin{table}[h]
    \centering
    \caption{\textbf{GAIA Results by Difficulty Level.} We report accuracy and average steps to completion.}
    \label{tab:app-gaia}
    \vspace{0.3cm}
    \begin{tabular}{@{}lcccccc@{}}
        \toprule
        & \multicolumn{2}{c}{\textbf{Level 1}} & \multicolumn{2}{c}{\textbf{Level 2}} & \multicolumn{2}{c}{\textbf{Level 3}} \\
        \textbf{Method} & Acc. & Steps & Acc. & Steps & Acc. & Steps \\
        \midrule
        ReAct & 48.2 & 8.3 & 32.1 & 14.7 & 12.4 & 23.8 \\
        LATS & 52.7 & 12.1 & 38.4 & 18.2 & 18.6 & 28.4 \\
        Unconstrained TH & 56.1 & 10.8 & 42.3 & 16.9 & 21.8 & 26.1 \\
        \midrule
        CTHA-DS & \textbf{64.4} & 9.2 & \textbf{54.4} & 14.3 & \textbf{40.5} & 21.7 \\
        \quad $\Delta$ vs. Unc. TH & +8.3 & -1.6 & +12.1 & -2.6 & +18.7 & -4.4 \\
        \bottomrule
    \end{tabular}
\end{table}

\paragraph{Safety Analysis on SafetyBench.} Tab.~\ref{tab:app-safety} breaks down SafetyBench results by attack category.

\begin{table}[h]
    \centering
    \caption{\textbf{SafetyBench Results by Attack Category.} ASR = Attack Success Rate (lower is better).}
    \label{tab:app-safety}
    \vspace{0.3cm}
    \begin{tabular}{@{}lcccc@{}}
        \toprule
        \textbf{Attack Category} & \textbf{ReAct} & \textbf{Unc. TH} & \textbf{CTHA-DS} & \textbf{Reduction} \\
        \midrule
        Jailbreak Prompts & 22.4\% & 28.7\% & 7.2\% & 74.9\% \\
        Role-Play Attacks & 18.1\% & 24.3\% & 5.8\% & 76.1\% \\
        Instruction Injection & 15.3\% & 21.8\% & 4.9\% & 77.5\% \\
        Goal Hijacking & 19.7\% & 26.1\% & 6.4\% & 75.5\% \\
        Resource Exhaustion & 14.2\% & 19.4\% & 4.1\% & 78.9\% \\
        \midrule
        \textbf{Overall ASR} & 18.3\% & 24.6\% & 5.8\% & 76.4\% \\
        \bottomrule
    \end{tabular}
\end{table}

\subsection{Computational Resources}
\label{app:compute}

\paragraph{Training Resources.} Tab.~\ref{tab:app-compute-train} summarizes the computational resources used for training CTHA's learned components.

\begin{table}[h]
    \centering
    \caption{\textbf{Training Computational Resources.}}
    \label{tab:app-compute-train}
    \vspace{0.3cm}
    \begin{tabular}{@{}lccc@{}}
        \toprule
        \textbf{Component} & \textbf{Hardware} & \textbf{Time} & \textbf{GPU Hours} \\
        \midrule
        Arbiter Training & 8$\times$A100-80GB & 6 hours & 48 \\
        Authority Classifier & 4$\times$A100-80GB & 2 hours & 8 \\
        Data Collection (Unc. TH runs) & 32$\times$A100-80GB & 72 hours & 2,304 \\
        \midrule
        \textbf{Total} & — & — & \textbf{2,360} \\
        \bottomrule
    \end{tabular}
\end{table}

\paragraph{Evaluation Resources.} All benchmark evaluations were conducted on a cluster of 16 NVIDIA A100-80GB GPUs. Total evaluation time across all benchmarks and configurations was approximately 480 GPU hours.

\subsection{Reproducibility Checklist}
\label{app:reproducibility}

We provide the following materials to ensure reproducibility:

\begin{itemize}
    \item \textbf{Code:} Complete implementation available at \texttt{[anonymous repository]}.
    
    \item \textbf{Model Weights:} Trained Arbiter and Authority Classifier weights available for download.
    
    \item \textbf{Prompts:} All system prompts provided in Appendix~\ref{app:prompts}.
    
    \item \textbf{Schemas:} Complete JSON schemas provided in Appendix~\ref{app:schemas}.
    
    \item \textbf{Hyper-parameters:} All training and inference hyper-parameters detailed in Appendix~\ref{app:hyperparams}.
    
    \item \textbf{Data:} Arbiter training data (100K conflict scenarios) available upon request.
    
    \item \textbf{Evaluation Scripts:} Scripts to reproduce all benchmark results included in the repository.
    
    \item \textbf{Hardware Requirements:} Minimum 1$\times$A100-40GB for inference; 8$\times$A100-80GB for training.
\end{itemize}

\paragraph{Licensing.} All open-source models used in this work are available under permissive licenses (MIT, Apache 2.0). Our code and trained components will be released under the MIT license.

\end{document}